\ifdefined\XeTeXversion\else\pdfoutput=1\fi
\PassOptionsToPackage{table}{xcolor}
\documentclass[10pt, logo, onecolumn, copyright]{nv}
\usepackage{graphicx}
\usepackage{tikz}
\usetikzlibrary{arrows.meta,positioning}
\definecolor{nvidiagreen}{HTML}{76B900}

\usepackage{mdframed}
\usepackage{color}
\usepackage{xcolor}
\usepackage{colortbl}
\usepackage[utf8]{inputenc}
\usepackage[T1]{fontenc}

\usepackage{amsfonts}
\usepackage{nicefrac}
\usepackage{microtype}
\usepackage{multirow}
\usepackage{multicol}
\usepackage{tabto}
\usepackage{xspace}
\usepackage{amsmath}
\usepackage{adjustbox}
\usepackage{enumitem}
\usepackage{wrapfig}
\usepackage{dblfloatfix}
\usepackage{times}
\usepackage{verbatim}
\usepackage{amssymb}
\usepackage{mathtools}
\usepackage{caption}
\usepackage{subcaption}
\usepackage{array}
\usepackage{colortbl}
\usepackage{booktabs}
\usepackage{bbm}
\usepackage{makecell}
\usepackage{float}
\usepackage{siunitx}
\usepackage{pifont}
\usepackage{marvosym}
\usepackage{listings}
\usepackage{pdflscape}
\usepackage{footmisc}
\usepackage{url}
\usepackage{tabularx}
\usepackage{hhline}
\usepackage{diagbox}
\usepackage{tcolorbox}
\usepackage[nameinlink]{cleveref}
\usepackage{hyperref}
\usepackage[square,sort,comma,numbers]{natbib}
\usepackage{fp}
\usepackage{authblk}
\usepackage{xspace}

\ifdefined\XeTeXversion
  \microtypesetup{tracking=false}
\fi

\crefname{section}{Sec.}{Sec.}
\crefname{equation}{Eq.}{Eqs.}
\crefname{figure}{Fig.}{Figs.}
\crefname{table}{Tab.}{Tabs.}
\crefname{appendix}{Appendix}{Appendices}
\title{Sol-H3: Recursive Self-Improvement for MiniMax-H3 Inference Acceleration on Sol-Engine across Cloud and Edge}

\correspondingauthor={\footnotesize \textnormal{\textsuperscript{$*$}Equal contribution. \quad NVIDIA Research, Efficient AI Team \& Singapore Lab. \\
}}

\author{
\parbox{\linewidth}{
\centering
\vspace{-5pt}
\normalfont\bfseries\fontsize{9.5pt}{13pt}\selectfont
Yitong Li\textsuperscript{$*$},
Jincheng Yu\textsuperscript{$*$},
Junsong Chen\textsuperscript{$*$},
Haopeng Li\textsuperscript{$*$},
Shuchen Xue,
\\
\normalfont\bfseries\fontsize{9.5pt}{13pt}\selectfont
Haozhe Liu,
Ping Luo,
Song Han,
Enze Xie
\vspace{2mm}
\\
\vspace{2pt}
}
}

\begin{abstract}
Video diffusion models are rapidly scaling and exhibiting enhanced generation capabilities. Among these recent advancements, \textbf{MiniMax-H3 stands out as a highly capable, production-level open-source model}. However, its 33-billion parameters and multi-step iterative denoising process introduce substantial computational overhead. Consequently, their practical production is hindered by generation latency in the cloud deployment like NVIDIA-GB200, alongside strict memory limits that pose further challenges at the edge device like DGX-Spark. To address these diverse hardware bottlenecks from cloud to edge device, we present a full-stack inference pipeline that integrates efficient algorithmic design with optimized operator implementations. Algorithmically, we introduce a \textbf{cross-resolution two-stage generation} scheduler that exploits the step-wise nature of diffusion: early low-resolution steps rapidly establish the global layout, while later high-resolution steps focus refinements of local and perceptual details. These stages are connected by a learned latent-to-latent mapping module, completely eliminating the computationally expensive VAE decode-reencode cycle for resolution transferring cross different resolutions. For operator implementation, we deploy a \textbf{Recursive Self-Improvement (RSI) loop} that searches kernel fusions and memory layouts, evaluating latency together with numerical agreement. These optimizations ensure our streamlined algorithmic design fully realizes its performance potential and practical latency benefits across varying hardware architectures. By combining these innovations, our pipeline achieves exceptional efficiency up to \textbf{30$\times$ speedup and 20\% memory saving}, delivering \textbf{up to 3.5$\times$ faster-than-real-time generation} on 8$\times$GB200 cloud nodes and \textbf{less than 1-min fully memory-resident execution} on DGX-Spark.
\end{abstract}

\begin{document}

\maketitle

\vspace{2pt}
\noindent\textbf{Links:}~{\hypersetup{urlcolor=nvidiagreen}\href{https://github.com/NVlabs/Sana/tree/sol-engine/models/minimax_h3}{Github Code} | \href{https://nvlabs.github.io/Sana/Sol-Engine/Sol-H3/}{Project Page}}

\begin{figure}[H]
\centering
\includegraphics[width=\linewidth]{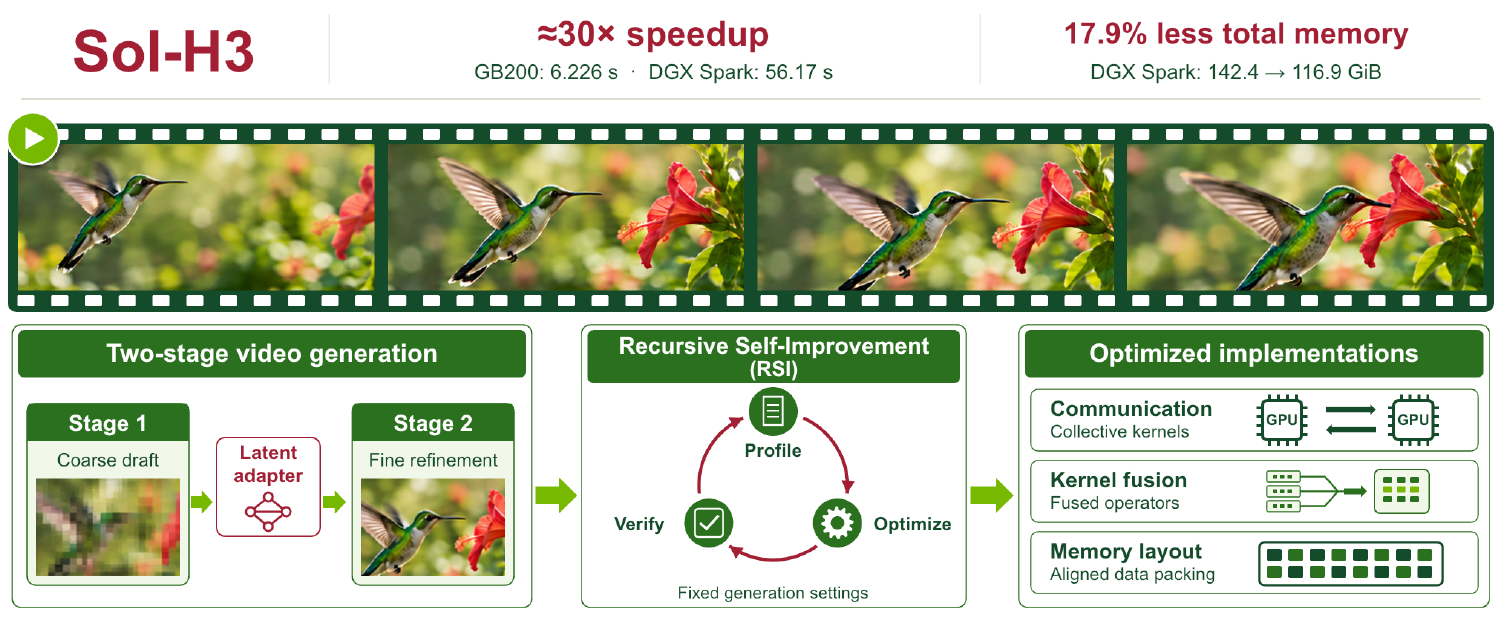}
\caption{\textbf{Sol-H3: two-stage video generation with recursive self-improvement.} A cross-resolution two-stage inference pipeline reduces the overhead of full-resolution denoising. Recursive Self-Improvement (RSI) profiles, optimizes, and verifies implementations under restricted generation settings, producing optimized kernels and effective memory layouts. The combination of this inference algorithm design and efficient operator implementation enables MiniMax-H3 to achieve approximately 30x acceleration and reduce GPU memory consumption by nearly 20\%.}
\label{fig:teaser}
\end{figure}

\clearpage

\section{Introduction}
\label{sec:intro}

Driven by model and data scaling, video diffusion models~\cite{ho2020ddpm,rombach2022ldm,peebles2023dit} have achieved remarkable generative prowess, with recent open-source releases like \textbf{MiniMax-H3 delivering production-grade quality}~\cite{minimax2026h3,kong2024hunyuanvideo,hunyuanvideo2025,wan2025,yang2025cogvideox,lightricks2026ltx25}. Despite these advances, the model's massive 33-billion parameter and the multi-step iterative denoising process~\cite{ho2020ddpm,lipman2023flowmatching} impose a severe computational burden. Furthermore, the specific bottlenecks encountered during inference vary drastically across different hardware platforms and deployment scenarios. In cloud environments, system throughput and generation cost act as the primary hurdles, ultimately determining the economic viability of the service. Conversely, edge deployments are tightly constrained by memory limits; exceeding the available HBM forces the system to cpu-offloading, precipitating a catastrophic collapse in inference latency. Since these hardware constraints, a single optimization strategy cannot solve both issues, making the overall optimization process much more difficult.

The acceleration of diffusion models has relied on two distinct paradigms. The first is algorithmic acceleration, which inherently alters the generation process through methods such as reducing sampling steps~\cite{salimans2022progressive,song2023consistency,luo2023lcm,yin2024dmd,sauer2023add}, caching denoising computation~\cite{liu2024timestep,zhou2025easycache,zhao2024pab,liu2025taylorseer,cachedit2025}, implementing sparse attention mechanisms~\cite{xi2025sparse,yang2025sparse,zhang2025vsa,zhang2025sta,li2025radial,xu2025xattention}, or modifying generation resolutions~\cite{ho2022cascaded,podell2023sdxl,jin2024pyramidal}, which are often in a lossy manner. The second is system-level acceleration, which preserves the mathematical output while applying low-level system optimizations, such as kernel fusion, memory layout restructuring, and communication scheduling~\cite{cutlassEpilogue,zhai2023bytetransformer,dao2022flashattention,guo2026coda}.

In this report, we optimize the inference pipeline from two perspectives of algorithm and implementation. \textbf{For algorithmic acceleration}, we leverage a fundamental evolutionary property of diffusion models: early denoising steps determine the overall structural layout of the video, while later steps synthesize high-frequency fine details. Building on this, we design a \textbf{cross-resolution two-stage generation pipeline} tailored to maximize efficiency at each phase. The first stage operates at a low resolution to rapidly establish the global layout and structural skeleton. The second stage shifts to a high resolution to focus entirely on refining visual details. To seamlessly connect these stages, we train a feature mapping module that performs direct latent-to-latent translation, completely eliminating the latency and memory overhead of the traditional pixel-level decode-and-reencode cycle. \textbf{For system-level implementation}, we introduce a \textbf{Recursive Self-Improvement (RSI) loop}~\cite{li2026sol} to conduct iterative and verifiable optimization. Operating under strict correctness verification, this RSI loop automatically explores a massive search space encompassing kernel fusions, memory layouts, and communication boundaries. The automated search consistently uncovers and optimizes low-level compute and communication bottlenecks, delivering massive kernel-level speedups across a wide range of hardware platforms. These optimization realizes the practical latency benefits of our efficient inference pipeline across varying hardware architectures, while requiring less human effort.

Integrating these algorithmic and systematic optimizations, we build a highly efficient full-stack inference pipeline across various hardware platform. On an 8$\times$GB200 node, Sol-H3 generates a five-second video with native audio in 1.434 seconds, approximately \textbf{3.5$\times$ faster than real-time playback} (\cref{tab:solh3_scaling}). On a single GB200, it achieves an end2end 6.23s generation, \textbf{22.2$\times$ speedup} over the serving baseline~\cite{h3super2026}. Furthermore, on a single DGX Spark, the entire pipeline remains fully HBM resident, completing the video generation in 56.2 seconds without triggering cpu offloading~\cite{solh3spark2026}, achieving a near 20\% memory reduction more than ${30\times}$ speedup .

We summarize the main contributions of this report below.

\begin{itemize}[leftmargin=1.5em, itemsep=2pt]
    \item \textbf{A cross-resolution two-stage generation pipeline}. Exploiting the step-wise nature of diffusion models, we allocates low-resolution earlier steps to global layout generation and high-resolution later steps to fine detail refinement, reducing the algorithmic redundancy of uniform-resolution denoising process.
    
    \item \textbf{A full-stack inference implementation for MiniMax-H3.} Through the co-design of algorithmic and system-level optimizations, our pipeline achieves an approximate 30$\times$ acceleration and a 20\% memory reduction across cloud and edge environments. This unlocks practical deployment for MiniMax-H3.
    
    \item \textbf{Automated system-level optimization via Recursive Self-Improvement (RSI).} We introduce an RSI loop that searches kernel fusions, memory layouts, collective layouts, and transport precision, profiles their latency, and checks numerical agreement. We distinguish bit-exact layout transformations from arithmetic fusions and approximate execution modes, giving the reported accelerations an explicit numerical scope.
\end{itemize}

\section{Related Work}
\label{sec:related}

\subsection{Video Diffusion Models} \label{subsec:related_video}

The current landscape of generative video is defined by open, large-scale diffusion transformers. Prominent base models include the CogVideo family \cite{hong2022cogvideo,yang2025cogvideox}, HunyuanVideo \cite{kong2024hunyuanvideo} and its successor HunyuanVideo 1.5 \cite{hunyuanvideo2025}, as well as Wan \cite{wan2025}, LongCat-Video \cite{meituan2025longcatvideo}, SANA-Video \cite{chen2025sana}, Cosmos 3 \cite{nvidia2026cosmos3super}, and JoyAI-Echo \cite{echo2026longvideo}. This report specifically serves MiniMax-H3 \cite{minimax2026h3}, a 33-billion parameter joint audio-video generator, alongside LTX-2.5 \cite{lightricks2026ltx25} and its predecessor LTX-2.3 \cite{lightricks2026ltx23}, which provide second-stage refinement. Our latent handoff combines spatial upsampling with a separate learned translation between the H3 and LTX VAE representations. The joint audio-video nature of MiniMax-H3 introduces a distinct serving challenge compared to vision-only models: the audio track must survive every optimization applied to the pipeline, not merely the visual frames.

To manage the computational complexity of high-resolution generation, cascaded and multi-resolution generation schedules are well-established recipes in the literature \cite{ho2022cascaded,jin2024pyramidal}. We employ a similar paradigm, utilizing a low-resolution draft from MiniMax-H3 followed by a high-resolution refinement pass with LTX-2.5. We emphasize that the low-resolution draft and high-resolution refine strategy is prior art; our contribution, detailed in \cref{subsec:twostage}, lies in the serving realization and the end-to-end execution stack rather than the conceptual schedule itself.

\subsection{Inference Acceleration for Video Diffusion} \label{subsec:related_acceleration}

Inference acceleration for diffusion transformers generally operates across three primary levers. The first lever is sparse attention, which mitigates the quadratic scaling of sequence length in video generation. Recent approaches include Sparse VideoGen \cite{xi2025sparse} and Sparse VideoGen2 \cite{yang2025sparse}, trainable sparse attention (VSA) \cite{zhang2025vsa}, Sliding Tile Attention \cite{zhang2025sta}, SpargeAttn \cite{zhang2025spargeattn}, XAttention \cite{xu2025xattention}, Radial Attention \cite{li2025radial}, DraftAttention \cite{shen2025draft}, and PISA \cite{li2026pisa}. Additionally, the SageAttention line of work \cite{zhang2025sageattention,zhang2025sageattention2,zhang2025sageattention3} provides highly optimized attention backends. We leverage similar principles for our sparse attention implementation, detailed in \cref{subsec:sparse}. The second lever involves caching mechanisms and techniques to reduce the number of effective steps, thereby limiting how much of the network is evaluated per step. Methods such as TeaCache \cite{liu2024timestep}, EasyCache \cite{zhou2025easycache}, Pyramid Attention Broadcast \cite{zhao2024pab}, TaylorSeer \cite{liu2025taylorseer}, and Cache-DiT \cite{cachedit2025} exploit temporal redundancy. These step-reduction techniques are orthogonal to our two-stage schedule; while caching reduces computation within a fixed resolution trajectory, our schedule alters the resolution trajectory itself.

The third lever encompasses quantization, token pruning, and kernel-level optimizations. Quantization techniques for diffusion models include ViDiT-Q \cite{zhao2025viditq}, PTQ4DiT \cite{wu2024ptq4dit}, Q-DiT \cite{chen2025qdit}, SVDQuant \cite{li2025svdquant}, AWQ \cite{lin2024awq}, and the adoption of microscaling formats \cite{ocp2023microscaling}. Token-level reduction strategies, such as ToMeSD \cite{bolya2023tomesd}, Astraea \cite{liu2025astraea}, TAPE \cite{li2026tape}, and CoReDiT \cite{li2026coredit}, further compress the computational graph. These optimizations are typically deployed within the broader context of modern serving systems and sequence parallelism frameworks, including vLLM \cite{kwon2023vllm}, SGLang \cite{zheng2024sglang}, DeepSpeed Ulysses \cite{jacobs2023ulysses}, and LongLive-2.0 \cite{chen2026longlive20}.

Throughout this report, we lean heavily on a fundamental distinction in inference optimization: methods that preserve the generation contract versus methods that alter the schedule or the step count. Techniques such as operator fusion (discussed in \cref{subsec:fusion}), exact attention backends, and lossless collective layouts provide like-for-like runtime speedups without changing the output. Conversely, schedule modifications produce a fundamentally different artifact. While these two categories of optimization compose effectively in practice, they must not be conflated when reporting speedups, as evaluating a different generation contract is not equivalent to accelerating a fixed one.

\subsection{Agentic and Self-Improving Optimization Systems} \label{subsec:related_agentic}

The general context for our recursive self-improvement loop is the rapid emergence of agentic coding and machine-learning experimentation systems. Frameworks such as SWE-agent \cite{yang2024sweagent}, AutoCodeRover \cite{zhang2024autocoderover}, Agentless \cite{xia2024agentless}, and OpenHands \cite{wang2024openhands} automate software engineering tasks, while benchmarks and systems like AgentBench \cite{liu2024agentbench}, MLAgentBench \cite{huang2024mlagentbench}, The AI Scientist \cite{lu2024aiscientist}, and AI Harness Engineering \cite{zhong2026ahe} explore autonomous research and experimentation. More specifically, this report belongs to the lineage of agents designed to write and optimize GPU kernels. The KernelBench benchmark \cite{ouyang2025kernelbench} demonstrated that frontier models perform this task poorly without iterative feedback. Subsequent systems have addressed this gap: CUDA-LLM \cite{chen2025cudallm} and CudaForge \cite{zhang2025cudaforge} focus on kernel generation, AccelOpt \cite{zhang2025accelopt} introduces a self-improving agentic system that curates an optimization memory of slow-fast kernel pairs, and AlphaEvolve \cite{novikov2025alphaevolve} employs an evolutionary coding agent for algorithmic discovery.

Our direct predecessor in this domain is the Sol video inference engine \cite{li2026sol}, an agent-native, full-stack acceleration framework. Sol composes caching, sparse attention, token pruning, quantization, and kernel fusion through a hierarchy of specialized skill agents and a central integrator agent. This report represents the deep application of the Sol framework to a single model family, extending the prior work by introducing a strict fail-closed contract as an explicit constraint on the optimization process.

This fail-closed constraint forms a critical positioning argument for our methodology. Unconstrained self-improving agents may inadvertently redefine the task they are optimizing---for instance, by silently reducing step counts or altering the numerical trajectory---which makes their reported performance gains difficult to attribute. Our recursive self-improvement loop is deliberately restricted: it searches realizations of a fixed generation contract (such as numerical formats, kernel boundaries, attention backends, collective layouts, and memory residency) but never alters the contract itself. Consequently, an accepted change is guaranteed to be a like-for-like reduction measured against the same baseline on the same hardware, and a rejected candidate yields a precise measurement rather than an unquantified judgment.

\section{Full-Stack Optimization for Cloud Deployments}
\label{sec:cloud}

Given the massive computational demands of video generation, cloud environments serve as the primary deployment vehicle. To maximize serving efficiency and reduce the compute cost per generation, we restructure the original single-stage, uniform-resolution denoising process into an asymmetric two-stage generation pipeline. We further apply a Recursive Self-Improvement (RSI) loop to iteratively optimize the execution stack. This RSI process systematically eliminates redundancy across each stage, elevating computational efficiency and generation throughput.

In this section, we detail algorithm, kernel, and system optimizations for the H3 runtime and its cross-resolution extension. The applicable configurations are evaluated separately in \cref{sec:experiments}. In particular, single-stage Sol-H3 on 8$\times$GB200 in \cref{sec:exp_solh3_scaling} generates a five-second video in 1.434 seconds, approximately 3.5$\times$ faster than real-time playback.

\subsection{Cross-Resolution Generation}
\label{subsec:twostage}

\paragraph{Cross-resolution two-stage generation.}

While the released pipeline runs the entire denoising trajectory at full resolution, we convert it into a two-stage efficient inference pipeline. Because early diffusion iterations establish global layout and motion while later steps synthesize high-frequency detail, executing the full trajectory at native resolution expends most compute resolving structural composition before fine features emerge. To address this, the Sol-H3 pipeline allocates spatial resolution dynamically, restructuring the conventional 49 denoising steps at full resolution ($1344\times768$) into an asymmetric multi-stage schedule:
\[
\underbrace{49 \text{ steps at } 1344{\times}768}_{\text{released pipeline}}
\quad\longrightarrow\quad
\underbrace{4 \text{ steps at } 672{\times}384}_{\text{Stage 1: content and motion}}
\;+\;
\underbrace{\text{latent handoff}}_{\times2\text{ upsample }+\text{ VAE translation}}
\;+\;
\underbrace{3 \text{ steps at } 1344{\times}768}_{\text{Stage 2: detail}} .
\]
Stage~1 generates core visual elements and the primary audio track using MiniMax-H3~\cite{minimax2026h3} augmented with the FastH3 VSA DataFree LoRA~\cite{zhang2025vsa}. It executes four update iterations at $672\times384\times124$ frames. Stage~2 synthesizes high-frequency detail using LTX-2.5~\cite{lightricks2026ltx25}. It executes three joint audio--video updates on the $1344\times768$ latent canvas using a resident BF16 LTX-2.5 dev backbone with distilled LoRA~450 applied at strength 0.8.

\begin{figure}[H]
\centering
\includegraphics[width=\linewidth,trim=0 85bp 0 110bp,clip]{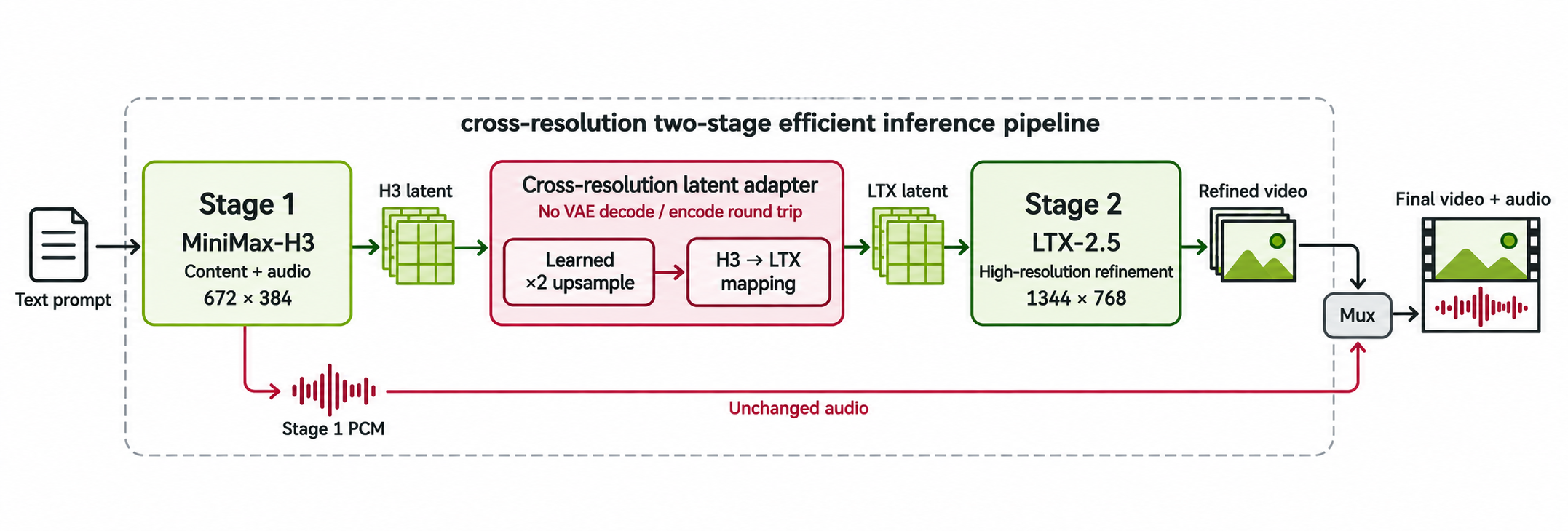}
\caption{\textbf{System overview.} We accelerates the generation process through a two-stage pipeline and a latent-to-latent connector. Stage~1 generates content at 672$\times$384 for 4-steps; an H3 latent upscaler doubles spatial resolution, and a separate adapter translates the result into LTX latent coordinates; Stage~2 refines at 1344$\times$768 for 3-steps. }
\label{fig:system_overview}
\label{fig:two_stage_pipeline}
\end{figure}

\paragraph{Cross-resolution latent space adapter.}
\label{subsec:adapter}

The two stages use independently trained VAEs~\cite{rombach2022ldm}, so matching spatial resolution alone does not make their latents interchangeable. H3 represents video with 24 channels and spatial compression of $16\times$, whereas the LTX-2.5 Conv Video VAE uses 128 channels and spatial compression of $32\times$. A direct connection would decode the H3 latent into pixels and re-encode those pixels with the LTX VAE, adding two large networks and their intermediate activations to the handoff. We replace this round trip with a learned latent-to-latent translator. Spatial enlargement remains a separate operation: the learned H3 $\times2$ upscaler first raises the draft resolution, and the translator then maps that representation to LTX at the same pixel resolution.

The main alignment difficulty is temporal. H3 encodes independent 17-frame chunks with nominal temporal compression of $4\times$, producing nonuniform token positions across chunk boundaries; LTX uses a causal $8\times$ temporal grid. We therefore align tokens by their physical frame positions rather than their tensor indices. A fixed front end concatenates a linearly interpolated H3 feature at each LTX time position with three slots containing the nearest-bin source tokens. Each source token is retained exactly once in the packed slots. A spatial pixel-unshuffle then moves each $2\times2$ neighborhood into channels, producing $24\times(1+3)\times4=384$ channels on the LTX grid. This parameter-free packing preserves the source features instead of averaging them away before learning the cross-VAE mapping.

On this aligned grid, the adapter combines a frozen $1\times1\times1$ affine skip, initialized by ridge regression, with a learned residual network. The network contains 22 blocks of width 752, each combining spatial convolution, depthwise temporal convolution, and a gated channel MLP. The complete translator has 194.76M parameters and predicts normalized 128-channel LTX latents.

Training pairs are obtained by encoding the same resized video with both frozen VAEs using their posterior modes. Let $z_H$ and $z_L$ denote the normalized H3 and LTX latents, $G$ the fixed alignment, and $A_\theta$ the adapter. We supervise both the latent prediction $\hat z_L=A_\theta(G(z_H))$ and its decoded video:
\begin{equation}
\mathcal{L}_{\mathrm{adapter}}
= \operatorname{MSE}(\hat z_L,z_L)
+\alpha\,\operatorname{MSE}\!\left(D_L(\hat z_L),D_L(z_L)\right),
\label{eq:adapter_loss}
\end{equation}
where $D_L$ includes inverse latent normalization, the frozen LTX Conv decoder, and conversion to RGB values clamped to $[0,1]$. Gradients pass through the decoder to the adapter while the VAE weights remain fixed. This decoded supervision accounts for the unequal visual effects of errors in different latent directions. The final training phases use 65,536 paired videos at $1344\times768$; implementation and evaluation details appear in \cref{app:adapter}.

At inference, the upscaler retains its prescribed input and output normalization, and the adapter output enters the refiner directly in normalized LTX coordinates. Applying LTX normalization a second time would change the input distribution. For the five-second serving configuration, temporal conversion produces 17 latent frames; the refiner consumes the first 16 to produce 121 output frames. The adapter operates only on video latents, leaving the Stage-1 audio unchanged. Removing the intermediate H3 decoder and LTX encoder reduces the handoff's compute and memory requirements, supporting the co-resident deployment described in \cref{sec:edge}.

\begin{figure}[H]
\centering
\begin{minipage}[c]{0.53\linewidth}
\centering
\begin{tikzpicture}[
  font=\footnotesize,
  box/.style={draw=black!45,rounded corners=2pt,align=center,
              minimum height=0.6cm,text width=2.35cm,inner sep=4pt},
  flow/.style={-{Stealth[length=1.7mm]},semithick},
  frozen/.style={box,fill=blue!7},
  trainable/.style={box,fill=nvidiagreen!15,draw=nvidiagreen!70!black}
]
\node[box,text width=3.1cm] (video) at (0,0) {Same resized source video};
\node[frozen] (h3) at (-1.65,-1.05) {Frozen H3 encoder};
\node[frozen] (ltx) at (1.65,-1.05) {Frozen LTX encoder};
\draw[flow] (video.south) -- ++(0,-0.15) -| (h3.north);
\draw[flow] (video.south) -- ++(0,-0.15) -| (ltx.north);
\node (zh) at (-1.65,-1.85) {$z_H$};
\node (zl) at (1.65,-1.85) {$z_L$};
\draw[flow] (h3) -- (zh);
\draw[flow] (ltx) -- (zl);
\node[trainable] (adapter) at (-1.65,-2.75) {Fixed packing $G$\\Residual adapter $A_\theta$};
\draw[flow] (zh) -- (adapter);
\node (pred) at (-1.65,-3.7) {$\hat z_L$};
\node[box,fill=orange!10] (latentloss) at (1.65,-3.7) {Latent MSE};
\draw[flow] (adapter) -- (pred);
\draw[flow] (pred) -- (latentloss);
\draw[flow] (zl) -- (latentloss);
\node[frozen,text width=4.85cm] (decodedloss) at (0,-4.85)
  {Frozen LTX decoder on both latents\\Decoded pixel MSE};
\draw[flow] (pred.south) -- (-1.65,-4.4) -- (decodedloss.north -| pred.south);
\draw[flow] (zl.east) -- (3.08,-1.85) |- (decodedloss.east);
\node[align=center,font=\scriptsize,text=black!70] at (0,-5.65)
  {Both losses update only the residual adapter.};
\end{tikzpicture}
\end{minipage}%
\hfill
\begin{minipage}[c]{0.44\linewidth}
\centering
{\small\textbf{H100 conversion profile}}\par\medskip
{\small
\setlength{\tabcolsep}{4pt}
\begin{tabular}{lrr}
\toprule
 & \textbf{ms} & \textbf{GiB} \\
\midrule
Full VAE & 12,748.6 & 13.788 \\
Tiny AutoEncoder & 201.9 & 23.846 \\
\textbf{Adapter} & \textbf{63.1} & \textbf{0.785} \\
\bottomrule
\end{tabular}}\par\smallskip
{\scriptsize BF16, batch 1, 192 frames, $1344\times768$.\\
Median latency and peak allocated memory.}\par\bigskip
{\small\textbf{Final held-out reconstruction}}\par\medskip
{\large\textcolor{nvidiagreen!70!black}{30.618\,dB\quad /\quad 0.8962}}\par\smallskip
{\scriptsize PSNR / SSIM on 256 test clips.\\
Compared with the decoded LTX teacher.}
\end{minipage}
\caption{\textbf{Cross-VAE latent translation.} Left: both frozen encoders process the same source video; latent and decoded-pixel supervision train the residual translator. The spatial upscaler is separate from this training task. Right: conversion cost for an earlier checkpoint and held-out quality for the final checkpoint of the same architecture. The profile excludes the upscaler and refiner (\cref{subsec:adapter_exp}).}
\label{fig:latent_adapter}
\end{figure}

\subsection{Communication Optimization}
\label{subsec:parallel}

We parallelize inference across GPUs, but communication and tensor layout transformations limit the realized speedup. Attention requires exchanges between sequence-sharded and head-sharded layouts, making the surrounding collectives a focus of the RSI optimization loop.

\paragraph{Optimized attention collectives.}
In a representative full-resolution H3 profile, layout transformations and collectives accumulated 94\,ms per denoising step, alongside 341\,ms of SDPA attention computation. The 94\,ms comprises 42\,ms of layout copies and 52\,ms of communication. The baseline executes three QKV collectives and one output collective per attention layer. Retaining packed QKV storage combines the three input exchanges into one, reducing the total to two collectives without changing the unquantized payload volume. A flat \texttt{(total, heads, head\_dim)} layout also removes the redundant batch dimension at batch size one. However, a naive PyTorch implementation achieved only $0.953\times$ the reference throughput, a 4.7\% reduction: \texttt{stack} followed by a destination-major \texttt{permute(...).contiguous()} copied the activation volume twice before communication. The optimized stride-aware kernels instead read Q, K, and V through their existing strides and write directly into the exchange buffer in one pass. The inverse head merge uses the same approach on the return path. These layout-only kernels are bit-identical to the reference permutations they replace.

\paragraph{Quantized communication.}
Under sequence parallelism~\cite{jacobs2023ulysses}, the packed QKV exchange and
attention-output return require two all-to-all collectives per attention layer.
The block-INT8 QKV format stores one token/head record as 384 quantized value
bytes, twelve group-32 finite-positive UE5M3 scale codes, and four zero padding
bytes, for 400 bytes in total. We evaluate two distinct output formats. The
block-INT8 format stores 128 value bytes and four scale codes, padded from
132 to 144 bytes to align each record to 16 bytes; this aligned format improves
the measured NCCL transfer performance on GB200 despite its larger payload.
The FP8 alternative instead transmits 128 raw E4M3 bytes per token/head,
without separate scale metadata or padding. All transmitted padding bytes are
explicitly initialized. Both quantized wire formats are approximate and are
validated separately from the bit-exact layout transformations.

\subsection{Compute Quantization}
\label{subsec:quant}

Attention and FFN linear projections process large token matrices and are important targets for low-precision GEMM acceleration~\cite{li2025svdquant,zhao2025viditq,chen2025qdit,wu2024ptq4dit,li2026fp4explorebf16train}. Their end-to-end benefit also depends on the cost of quantization, scale layout conversion, and surrounding data movement. We therefore evaluate the arithmetic kernels together with their producers and communication interfaces.

\paragraph{MXFP8 GEMM computation.}
Sol-H3 uses MXFP8 GEMMs for attention and FFN projections in zero-indexed
transformer blocks 2--46, inclusive, on the validated SM100-family path~\cite{ocp2023microscaling}. Weights and activations use
E4M3 values with one E8M0 scale for each group of 32 values along the reduction
dimension. The scale buffers use the cuBLASLt-compatible
\texttt{SWIZZLE\_32\_4\_4} layout. Fused RMSNorm/modulation and SwiGLU producers
write the E4M3 activations and swizzled scales directly, eliminating the
intermediate BF16 tensor and a separate quantization pass. Boundary blocks,
AdaLN projections, refiners, VAE, and text/audio encoders retain BF16
computation. The quantization is approximate.

\paragraph{Joint optimization with quantized communication.}
Sol-H3 reuses the raw E4M3 values returned by attention as inputs to the
quantized output projections, supplying unit E8M0 scales. The receiver merges
the rank-major heads directly into the projection's row-major input, avoiding
an intermediate BF16 tensor and subsequent dynamic quantization. Integrating
this reuse with the fused activation producers removes redundant data
conversions between attention, communication, and linear computation.

\subsection{Sparse Attention}
\label{subsec:sparse}

\paragraph{Per-step dynamic sparsity.}
In the refinement stage, the token count grows substantially with resolution, sharply increasing the computational cost of attention~\cite{dao2022flashattention,dao2024flashattention2}.
We therefore employ Sol-Attn~\cite{li2026solattn}, a training-free block-sparse attention method~\cite{xi2025sparse,yang2025sparse,zhang2025spargeattn,li2026pisa} that uses online thresholding to select a small subset KV blocks for computation.
Specifically, for each query block, the threshold is set to $\mu + \tau\sigma$, where $\mu$ and $\sigma$ denote the mean and standard deviation of its block-level attention scores across KV blocks, and $\tau$ is a tunable hyperparameter that controls sparsity.
Only KV blocks whose scores exceed this threshold are selected for attention computation; a larger $\tau$ raises the threshold and thus retains fewer blocks.
We observe that sparsification has a greater impact on global image coherence during early, high-noise refinement steps than during later, low-noise steps.
We therefore adopt a step-dependent schedule that progressively increases sparsity, setting $\tau$ to $1.0$, $1.25$, and $1.5$ for the three refinement steps, respectively.
To accommodate different GPU architectures, we use cuDNN block-sparse attention (BSA) and a custom CuTe DSL kernel~\cite{cutlassEpilogue} as alternative backends, improving kernel efficiency and multi-GPU serving performance.

\paragraph{Safety rails and the exact prefix KV sink.}
In practice, we keep the first two transformer layers dense throughout denoising, as errors introduced while global structure is still forming may persist through subsequent layers.
Within the remaining attention layers, we use distinct attention paths for prefix and target-video queries.
MiniMax-H3 packs each sequence as $[\,\texttt{text}\mid\texttt{conditioning video}\mid\texttt{audio}\mid\texttt{target video}\,]$, so all tokens preceding the target video form a contiguous prefix.
Queries from this prefix attend to all KV tokens through dense cross-attention.
The remaining target-video queries use block-sparse attention, with the prefix KV blocks serving as always-attended sinks that are evaluated exactly regardless of the routing threshold.
Importantly, this protection covers the entire prefix rather than text tokens alone: the audio tokens are themselves generated, with the model predicting audio velocities at these positions.
In validation, we observed severe dialogue degradation even in a case that achieved the highest visual-quality score among the compared variants, highlighting the need to preserve audio fidelity alongside visual quality.
This design preserves dense connectivity between the multimodal prefix and the full sequence while retaining block-sparse attention among target-video tokens.

\subsection{Kernel Fusion and Precision Preservation}
\label{subsec:fusion}

The DiT~\cite{peebles2023dit} repeatedly writes wide intermediate tensors to HBM and reads them back for subsequent elementwise operations. Fusion reduces this traffic by consuming intermediates in registers. H3 combines RMSNorm, partial rotary embeddings, SwiGLU, and per-row indexed modulation, requiring adaptations to fusion kernels designed for other architectures. The RSI loop searches implementation and launch configurations, profiles their latency, and checks their numerical behavior against the corresponding reference.

\paragraph{Kernel Fusion.}
The RSI loop identifies three operator groups across the 50 H3 transformer
blocks: (1) residual addition with indexed gating, RMSNorm, and indexed
scale/shift modulation; (2) QK normalization with partial RoPE; and (3) SwiGLU
splitting, SiLU, and multiplication. At a representative four-rank
context-parallel shape of 9562 local rows and hidden width 5376, these fusions
reduce repeated reads and writes of wide intermediate tensors. Pure layout
transformations are bit-identical to their reference permutations. Arithmetic
fusions preserve the operator structure, but changes in intermediate precision
and floating-point rounding can prevent bitwise equivalence to the original
eager implementation; we distinguish these from the separately validated
LoRA consumer-fusion equivalence below.

\paragraph{Precision-preserving LoRA fusion.}
Weight merging removes LoRA's extra GEMMs, but can change a few-step adapter's numerical behavior. Small updates can round back to the original weights when $W+BA$ is stored in BF16, even if the merge itself is computed in FP32. The real-arithmetic identity $(W+BA)x=Wx+B(Ax)$ therefore does not imply equivalence in BF16. In a controlled diagnostic of the evaluated eight-step adapter, 86.1--94.4\% of nonzero weight updates in six inspected projections rounded back to their original values. These are per-projection weight statistics, not a model-wide fraction or a perceptual-quality loss; \cref{app:lora_fusion} gives the diagnostic controls.

We instead retain the base and LoRA branches and fuse their addition into the next consumer kernel. Each projection still executes the same three GEMMs with unchanged weights and reduction order. The fused consumer explicitly rounds $\mathrm{base}+\delta$ to BF16 before QK normalization, RoPE and packing, residual modulation, or SwiGLU; the final FFN consumer also retains the intermediate BF16 gate product. Preserving these rounding points removes the wide sum's HBM write/read pair while retaining the separate-branch arithmetic. The implementation specializes to one active adapter at unit scale with BF16 linear compute, independent of the adapter's name. Its numerical agreement and runtime are evaluated against native branches with the other acceleration settings held constant in \cref{sec:exp_lora_qualitative}.

\subsection{VAE Decoding}
\label{subsec:vae}

Latent diffusion requires decoding to pixels~\cite{rombach2022ldm}. The optimizations in this subsection concern the H3 video decoder used in
single-stage H3-output paths. They are separate from the latent handoff in
\cref{subsec:adapter}, which bypasses H3 decoding and passes video latents to
the LTX refinement and decoding path.

\paragraph{Parallel decoding.}
Context parallelism shards the transformer stack but does not by itself
partition video decoding. In the evaluated H3 configuration, each rank would
otherwise decode the full video. At $1344\times768$ and 124 output frames, the
H3 decoder processes seven temporal clips with 28 equal-sized spatial tiles
per clip, for 196 tile decodes. Distributing these independent tiles over eight
ranks reduces the recorded decode latency from 7.55\,s to approximately
1.16\,s while retaining the original stitch order.

\paragraph{Local tile batching and compilation.}
Sharding leaves four tile slots per rank in each temporal clip. Decoding these
four tiles as one local batch, rather than launching the decoder separately
for each tile, reduces latency from 1.1597\,s to 0.6021\,s, a $1.93\times$
speedup in the recorded configuration. Compiling the batched decoder reaches
0.3593\,s, or $3.23\times$ relative to the 1.1597\,s sharded baseline, and
reduces the measured decoder peak memory from 18.8\,GB to 15.6\,GB. These are
decode-level measurements rather than whole-pipeline speedups. Compilation is
not bit-identical to eager decoding; the recorded maximum absolute deviation
is approximately 0.021.

\paragraph{Global tile batching.}
A separate optimization combines tiles across all seven temporal clips. The
196 real tiles then occupy 200 execution slots, or 25 per rank, with only four
padded duplicates, instead of 224 slots and 28 duplicates for per-clip
scheduling. The global path invokes the batched decoder and gather once rather
than seven times and restores the original tile and blend order afterward.
Changing the batch shape can change the selected GEMM implementation: the
recorded comparison against the per-clip path has maximum absolute deviation
0.013672 and decoder-output PSNR 75.75\,dB over the $[-1,1]$ range. These values
characterize that test, rather than a general error bound or perceptual-quality
guarantee. The measurements and implementation discussed here use eight
ranks; they do not establish the same batching benefit on a single GPU.

\section{Full-Stack Optimization for Edge Deployments}
\label{sec:edge}

While cloud environments serve as the primary deployment vehicle due to massive computational demands, there is substantial demand for executing these generative models locally or on edge devices. To enable efficient local deployment, we undertake a comprehensive set of optimizations targeting strict reductions in both latency and device memory footprint. The binding constraint in this regime is device memory residency. It requires the entire working set, including the text encoder, drafting model, refiner, and VAE, must remain simultaneously resident in HBM. If the working set does not fit, the runtime must offload weights to host memory and stream them back on demand, incurring a substantial host-device communication cost that dominates request latency.

On a single DGX Spark, where the CPU and GPU share a 119.68\,GiB unified memory pool, our optimizations represent the critical difference between a viable deployment and an out-of-memory failure. Specifically, a naive two-stage deployment demands 142.4\,GiB, exceeding the hardware capacity by 22.7\,GiB. Our optimized configuration consumes only 116.9\,GiB, preserving a 2.8\,GiB safety margin during sampling.

In this section, we detail the specific optimization techniques applied to achieve these latency and memory objectives, focusing on strategies such as deterministic modulation caching, prompt caching, and weight quantization. Ultimately, deploying this optimized pipeline on a single-accelerator edge environment, such as a GB10, successfully fits the complete generative stack onto the device while delivering highly responsive local inference.

\subsection{AdaLN precompute}
\label{subsec:cache}
For MiniMax-H3, approximately 13B of the model's 33B parameters reside in the \texttt{adaln\_proj} module. At inference time, this module recomputes the modulation table at every denoising step, amounting to 2500 recomputations per generated video. Because the projection's only input is the timestep embedding, and that embedding depends solely on the sampling schedule fixed before the denoising loop, every one of those recomputations produces a value that was already knowable in advance. Consequently, this repetition is pure redundancy, with the module imposing waste along two dimensions: device memory is consumed because the wide projection weights must stay resident, and compute and bandwidth are expended regenerating the same table at each step. Our runtime precomputes the entire table once and maintains a cache indexed by (block, step). This eliminates two severe bottlenecks. First, it saves about 24\,GB of device memory, as replacing the 26\,GB of \texttt{adaln\_proj} weights with a precomputed table of $\approx 1.5$\,GB lowers denoiser residency from 61.7\,GB to $\approx 37$\,GB. Second, it eliminates about 26\,GB of HBM reads per step; the reference pipeline streams 520\,MB of weight data across HBM per block to produce just nine output rows, incurring a bandwidth tax with minimal arithmetic intensity.

\subsection{Prompt caching}
\label{subsec:promptcache}
As established in \cref{subsec:twostage}, Stage 1 has already fixed the global layout, semantic content, and motion of the generated video. The role of Stage 2 is mainly local high-frequency refinement. Therefore, Stage 2 does not require per-sample text conditioning: the sample-specific information it needs is already carried by the Stage-1 latent it receives, and a generic, prompt-independent conditioning signal suffices.

Instead of encoding the user's prompt dynamically for Stage 2, we evaluate a single canonical generic refinement prompt once and offline, \textit{4K, refined, high quality, cinematic detail, clean textures, natural motion}. This prompt is processed through the INT8 Gemma text encoder and the INT8 dev connector, and we retain only the compact post-connector video and audio context tensors to feed to Stage 2 as a cache. Consequently, neither the Stage-2 text encoder nor its connector is ever loaded during online inference, freeing the 16.2\,GiB for other modules. 

\subsection{Weight quantization}
\label{subsec:memquant}
For deployment on a single DGX Spark (GB10), quantization is essential to keeping the two-stage pipeline resident within the device's 119.68\,GiB CPU--GPU unified memory pool. The edge configuration pairs an FP8 H3 draft DiT~\cite{ocp2023microscaling} with an NVFP4 AWQ Qwen prompt encoder~\cite{lin2024awq,ocp2023microscaling,li2026fp4explorebf16train,solh3spark2026}, with reported resident footprints of 23.4\,GiB and 14.6\,GiB, respectively. The runtime reads each layer's quantization configuration from the checkpoint, allowing
different projections to use different precision settings. Qwen encodes each incoming prompt during online inference. The INT8 Gemma encoder and connector are used only offline to construct the cached Stage-2 conditioning described in
\cref{subsec:promptcache}, and neither remains resident during request processing.

\section{Caching Reference Tokens for Reference to Video Generation}
\label{sec:ref2av}
\label{subsec:refcache}

In many practical deployment scenarios, generation is conditioned on user-provided reference materials. In this reference-to-video and audio (ref2av) setting, the model synthesizes output video and audio based on a set of conditioning reference images or videos. As the volume of reference materials data grows, the interaction between reference tokens and generation tokens becomes a struggler of generation latency. Thus, we propose a reference token caching strategy to further reduce latency in this setting~\cite{liu2024timestep,zhou2025easycache,zhao2024pab}.

\textbf{Reference computation redundancy.} During the standard diffusion inference process, the model executes $N$ denoising steps. In a naive implementation, every self-attention evaluation across these $N$ steps recomputes the reference tokens in full, with no computations skipped. However, unlike the target video tokens which are iteratively denoised, the reference latent fed into the Diffusion Transformer (DiT)~\cite{peebles2023dit} is a clean latent. Because no noise is added to the reference inputs, their representations vary only weakly with the timestep embedding~\cite{liu2024timestep,liu2025taylorseer}. Consequently, recomputing the full self-attention for the reference tokens at every denoising step introduces potential computational redundancy.

\textbf{Generation with heavy conditioning.} This redundancy becomes a severe bottleneck when processing extensive reference inputs. A single generation request may supply up to 9 reference images or 3 reference videos. Under these heavy conditioning workloads, the number of reference tokens can easily exceed the number of generation tokens themselves. At this scale, the redundant recomputation of reference tokens ceases to be a marginal overhead and instead dominates the overall inference cost, leading to a significant waste of computational resources.

\paragraph{Reference KV caching mechanism}. To eliminate this inefficiency, we introduce a reference KV caching mechanism. The core principle is to evaluate the reference branch in full only during the first DiT step. During this initial step, the key (K) and value (V) tensors for the reference tokens are computed and cached. For all subsequent denoising steps, the reference tokens are no longer recomputed; instead, their cached KV tensors are retained and reused unchanged (the cached KV is never iterated or refreshed). The subsequent steps only evaluate the generated tokens, which attend to the cached reference KV. Because the first step remains a full computation, the conditioning signal is established exactly, and only the redundant repetitions in later steps are bypassed.

We note that this caching strategy is an approximation rather than a bitwise-identical transformation. Although the input reference latent is clean, its intermediate representations naturally drift slightly across steps due to the changing timestep embeddings in the DiT blocks. However, from the qualitative comparison, we find that the faithful recomputation pipeline and caching pipeline show minor deviation, demonstrating that the caching strategy is a practical approximation method for reference to video generation especially when reference material is heavy. 

\begin{figure}[H]
    \centering
    \includegraphics[width=0.78\linewidth]{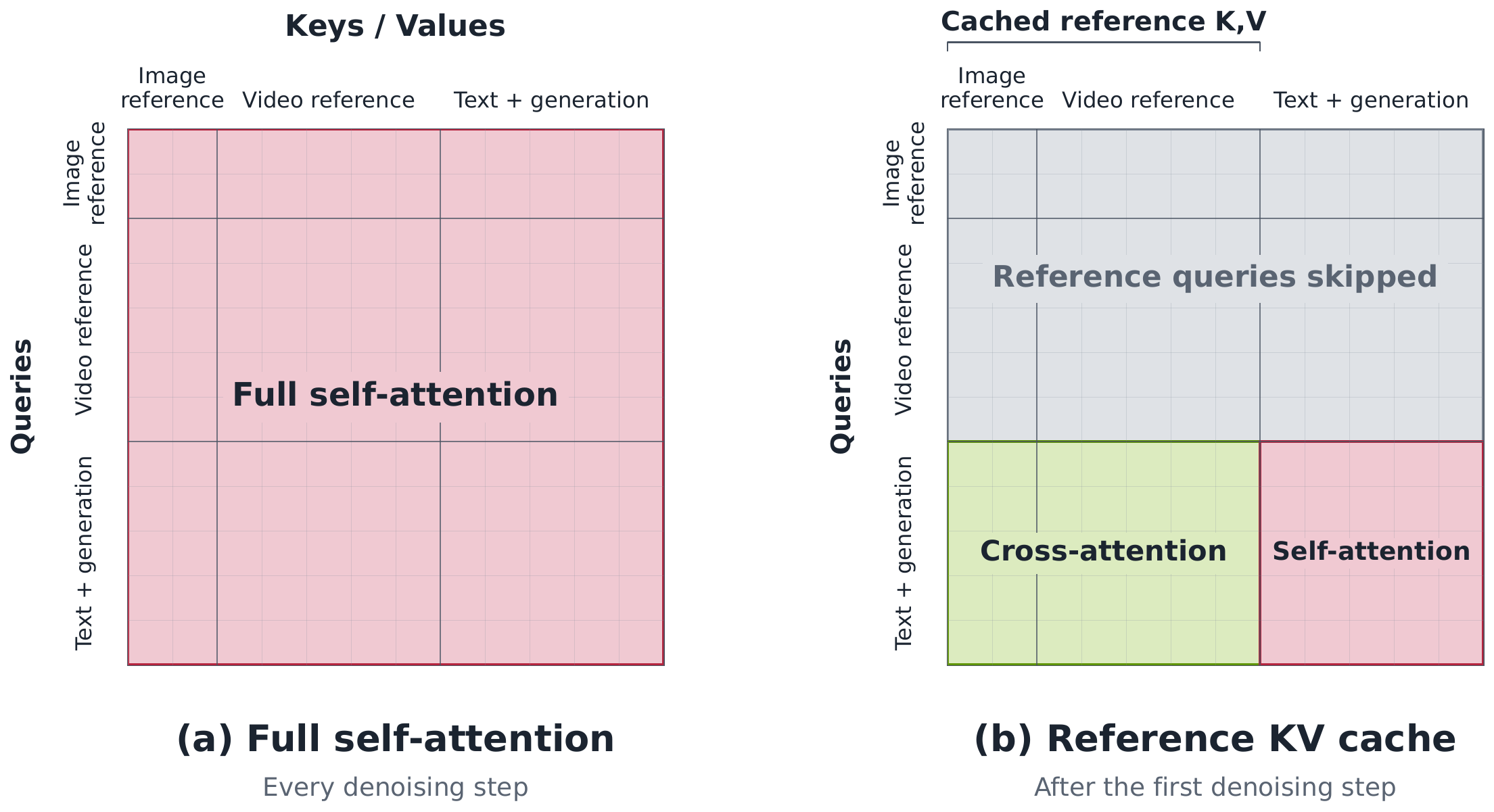}
    \caption{\textbf{Attention computation with reference KV cache.} Left: full self-attention over all tokens, the native inference mode of MiniMax-H3~\cite{minimax2026h3}. Right: after the first denoising step, image and video reference query rows are skipped (gray). Text and generation queries attend to cached reference K,V (cross-attention) and to each other (self-attention).}
    \label{fig:refcache_sweep}
\end{figure}

\section{Experiments}
\label{sec:experiments}

\subsection{Cross-Resolution Two-Stage Acceleration}
\label{subsec:twostage_exp}

We evaluate the two-stage cross-resolution pipeline results measured on a single NVIDIA GB200. For a 5~s workload at 1344$\times$768 resolution, the pipeline achieves an end-to-end latency of 6.852~s, comprising 4.173~s for Stage 1 and 2.679~s for Stage 2. This represents a 22.2$\times$ speedup over the SGLang~\cite{zheng2024sglang} baseline of 152.3~s~\cite{h3super2026}. For a 10~s workload at the same 1344$\times$768 resolution, the end-to-end latency is 14.931~s (9.820~s for Stage 1 and 5.111~s for Stage 2), yielding a 27.7$\times$ speedup over the SGLang baseline of 414.1~s.

\begin{figure}[!t]
\centering
\includegraphics[width=\linewidth]{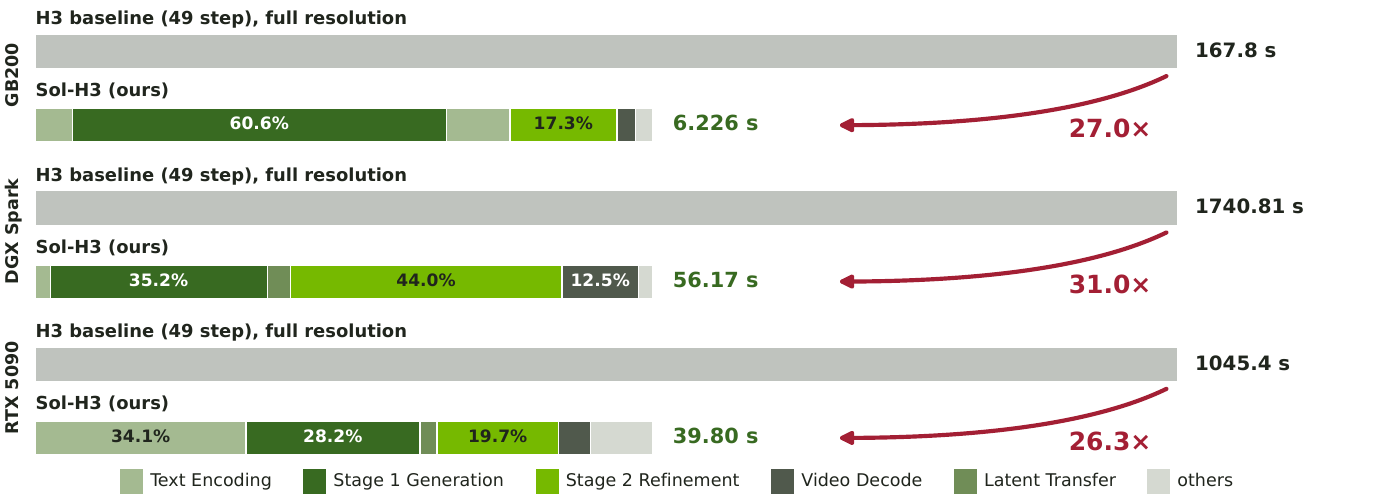}
\caption{\textbf{Latency profiles of Sol-H3 inference.} 
The same 5\,s $1344\times768$ workload on one GB200 (top), one DGX~Spark (middle) and one RTX~5090 (bottom), each against a full-resolution 49-step MiniMax-H3 baseline~\cite{h3super2026,solh3spark2026}. 
The optimized request completes in 6.226\,s on GB200, 56.17\,s on Spark and 39.80\,s on RTX~5090, a speedup of more than 20$\times$ in every case. The profiles differ in where the time goes: Stage-1 generation dominates on GB200 at 60.6\%, Stage-2 refinement dominates on Spark at 44.0\%, and on RTX~5090 prompt encoding is the largest single block at 34.1\%, since Qwen is loaded and released per request under CPU offload.}
\label{fig:latency_profiles}
\end{figure}

\begin{figure}[!t]
\centering
\scriptsize
\setlength{\tabcolsep}{2pt}
\renewcommand{\arraystretch}{1.0}
\begin{tabular}{@{}cc@{}}
\multicolumn{2}{@{}l@{}}{\parbox{\linewidth}{\raggedright\textbf{$1344\times768$, 5\,s} --- \emph{A man and woman exchange a tense remark on a crowded Japanese street.}}} \\[2pt]
\includegraphics[width=0.49\linewidth]{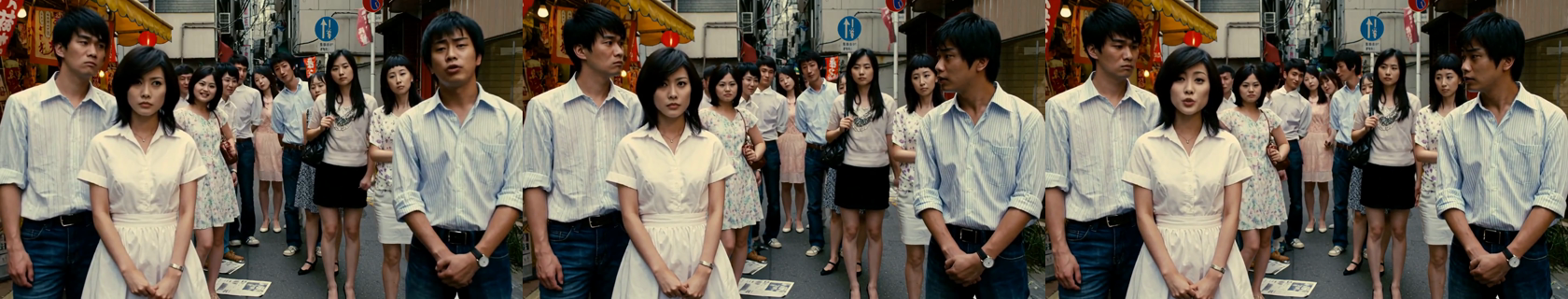} &
\includegraphics[width=0.49\linewidth]{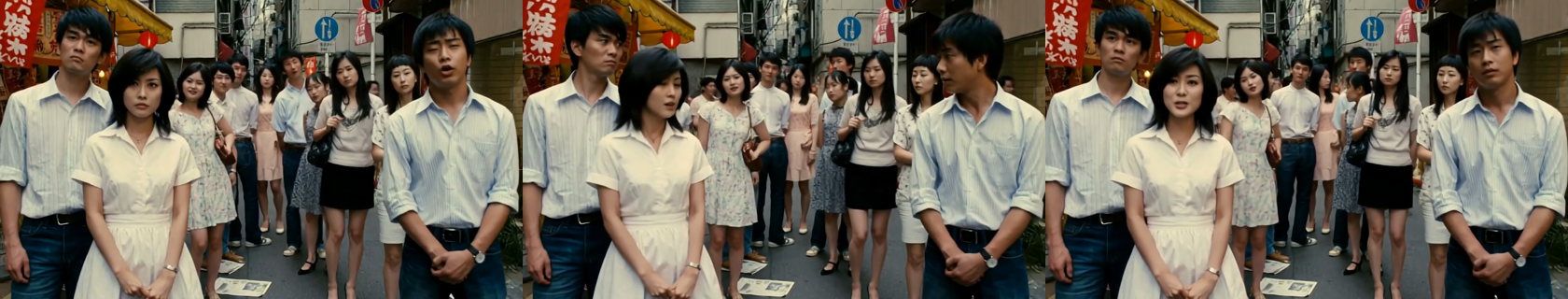} \\[1pt]
Baseline, 50 steps & \textbf{Sol-H3 (ours)} --- \textbf{\textcolor{nvidiagreen}{22.2$\times$}} \\[5pt]
\multicolumn{2}{@{}l@{}}{\parbox{\linewidth}{\raggedright\textbf{$1344\times768$, 10\,s} --- \emph{A man speaks to a handheld camera on a quiet suburban street.}}} \\[2pt]
\includegraphics[width=0.49\linewidth]{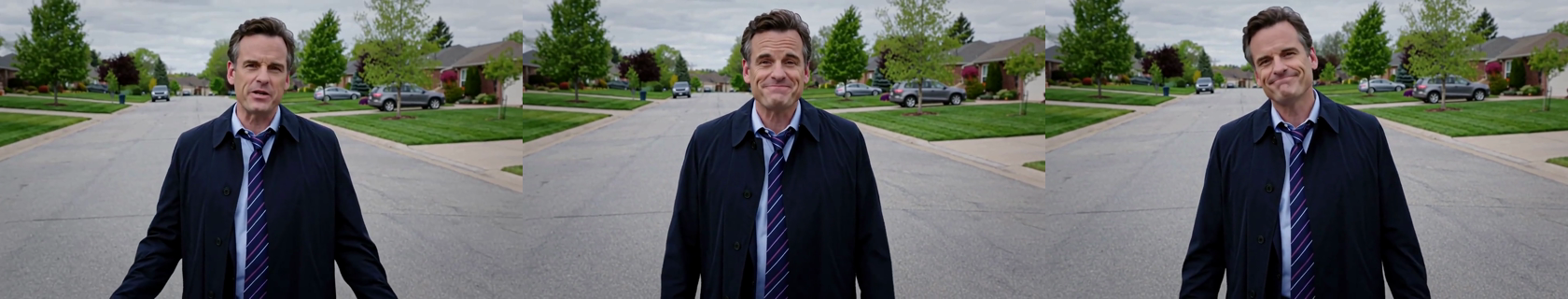} &
\includegraphics[width=0.49\linewidth]{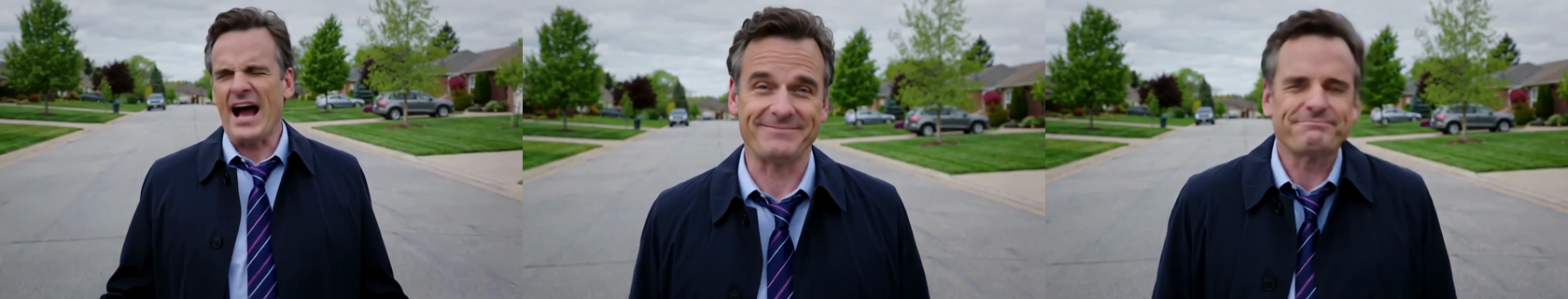} \\[1pt]
Baseline, 50 steps & \textbf{Sol-H3 (ours)} --- \textbf{\textcolor{nvidiagreen}{27.7$\times$}} \\[5pt]
\multicolumn{2}{@{}l@{}}{\parbox{\linewidth}{\raggedright\textbf{1080p, 5\,s} --- \emph{Two martial artists face one another in a bamboo forest.}}} \\[2pt]
\includegraphics[width=0.49\linewidth]{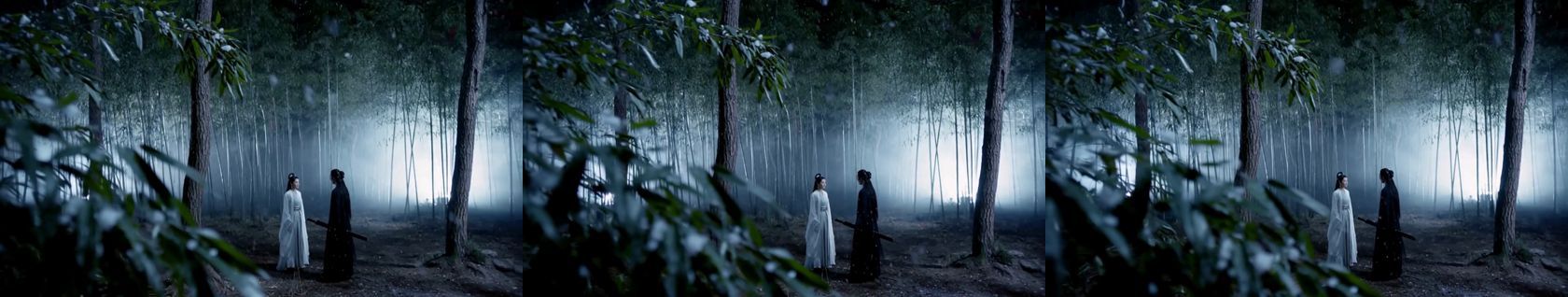} &
\includegraphics[width=0.49\linewidth]{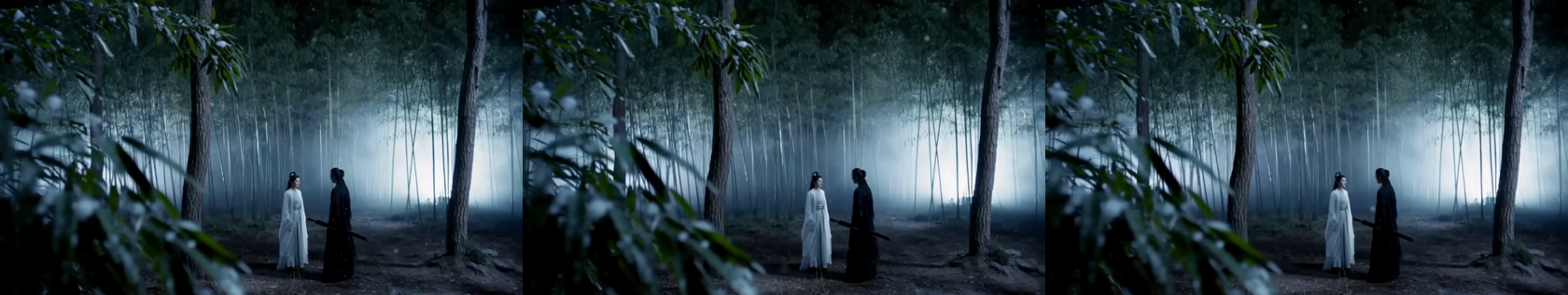} \\[1pt]
Baseline, 50 steps & \textbf{Sol-H3 (ours)} --- \textbf{\textcolor{nvidiagreen}{42.6$\times$}} \\
\end{tabular}
\caption{\textbf{Qualitative comparison between Sol-H3 and the baseline.} Sol-H3(right) drafts at a lower resolution and refines back to the target resolution. While preserving scene structure, subject identity and motion continuity with minor visual differences, Sol-H3 delivers a significant speedup of more than \textbf{20$\times$}~\cite{h3super2026}.}
\label{fig:twostage_qual}
\end{figure}

For edge device, the single DGX Spark deployment~\cite{solh3spark2026} reaches 56.17~s end to end for the same 5~s 1344$\times$768 workload, yielding a 31$\times$ speedup over the baseline of roughly 1740.81~s. The latency share breakdown for this deployment is as follows (\cref{fig:latency_profiles}): LTX 3-step refinement accounts for 44.0\%, H3 4-step generation for 35.2\%, LTX VAE decode for 12.5\%, H3 upscaler and VAE adapter for 3.8\%, Qwen prompt encoding for 2.4\%. A qualitative comparison of the generated frames is provided in \cref{fig:twostage_qual}.

The same two-stage pipeline also runs on a single consumer GPU. On one RTX~5090 the 5~s $1344\times768$ request completes in 39.80~s, against a 1045.40~s full-resolution 49-step baseline measured on the same card, a 26.3$\times$ speedup. The latency composition differs from both datacenter and DGX~Spark deployments: Qwen prompt encoding is the single largest block at 34.1\%, ahead of H3 4-step generation at 28.2\% and LTX 3-step refinement at 19.7\%, with LTX VAE decode at 5.1\% and the upscaler and latent adapter at 2.8\%. The text encoder dominates not because encoding is expensive but because the encoder is rebuilt for every request: of the 13.568\,s attributed to this block, the NVFP4 AWQ Qwen forward pass accounts for 0.518\,s and the rest is loading and teardown~\cite{lin2024awq}. Unlike a one-time startup cost, this is charged on each request, because the 5090 cannot hold the encoder in GPU HBM together with both generation stages and therefore runs under CPU offload. Denoising itself is therefore no longer the limiting term on consumer hardware; prompt-encoder residency is, which is the same pressure the Stage-2 conditioning cache in \cref{subsec:promptcache} relieves on DGX~Spark.

\subsection{Sol-H3 Latency on GB200 GPUs}
\label{sec:exp_solh3_scaling}

We evaluate Sol-H3 on one, four, and eight NVIDIA GB200 GPUs~\cite{solh32026,nvidia2026hgxcomponents} at 1344$\times$768 resolution and 24 FPS, generating reference-free video with stereo audio. All nine combinations of GPU count and output duration use the same prompt, seed 20260903, and merged four-step LoRA~\cite{hu2022lora,salimans2022progressive,song2023consistency,yin2024dmd}. The evaluated system combines SOL/BSA attention, quantized QKV and attention-output communication, MXFP8 GEMMs with fused activation producers and output reuse, and the VAE optimizations described in \cref{sec:cloud}.

\begingroup
\definecolor{scalingdark}{HTML}{386A21}
\definecolor{scalingwash}{HTML}{EFF6E5}
\newcommand{\scalingbest}[1]{\textbf{\textcolor{scalingdark}{#1}}}
\newcommand{\scalinggain}[1]{\textbf{\textcolor{scalingdark}{#1$\times$}}}
\begin{table}[H]
\centering
\caption{\textbf{Sol-H3 latency on GB200s.} Baseline of MiniMax-H3 and Sol-H3 latency for 5, 10, and 15 s outputs on NVIDIA GB200 GPUs. It delivers more significant speedups with longer duration of generated videos.}
\label{tab:solh3_scaling}
\begingroup
\fontsize{8.6}{11}\selectfont
\setlength{\tabcolsep}{2.7pt}
\renewcommand{\arraystretch}{1.48}
\begin{tabular*}{\linewidth}{@{\extracolsep{\fill}}crrr@{\hspace{8pt}}rrr@{\hspace{8pt}}rrr@{\hspace{\tabcolsep}}}
\toprule
\multirow{2}{*}{\textbf{GPUs}} & \multicolumn{3}{c}{\textbf{5 s output}} & \multicolumn{3}{c}{\textbf{10 s output}} & \multicolumn{3}{c}{\textbf{15 s output}} \\
\cmidrule(lr){2-4}\cmidrule(lr){5-7}\cmidrule(lr){8-10}
 & Base H3 & \scalingbest{Sol-H3} & Speedup & Base H3 & \scalingbest{Sol-H3} & Speedup & Base H3 & \scalingbest{Sol-H3} & Speedup \\
\midrule
\textbf{1} & 129.898 & \scalingbest{9.475} & \cellcolor{scalingwash}\scalinggain{13.71} & 376.942 & \scalingbest{24.309} & \cellcolor{scalingwash}\scalinggain{15.51} & 746.885 & \scalingbest{44.958} & \cellcolor{scalingwash}\scalinggain{16.61} \\
\textbf{4} & 35.328 & \scalingbest{2.530} & \cellcolor{scalingwash}\scalinggain{13.97} & 100.440 & \scalingbest{6.212} & \cellcolor{scalingwash}\scalinggain{16.17} & 194.930 & \scalingbest{11.471} & \cellcolor{scalingwash}\scalinggain{16.99} \\
\textbf{8} & 18.250 & \scalingbest{1.434} & \cellcolor{scalingwash}\scalinggain{12.73} & 50.660 & \scalingbest{3.350} & \cellcolor{scalingwash}\scalinggain{15.12} & 99.513 & \scalingbest{6.062} & \cellcolor{scalingwash}\scalinggain{16.41} \\
\bottomrule
\end{tabular*}
\endgroup
\par\vspace{4pt}
{\footnotesize NVIDIA GB200, 1344$\times$768, 24 FPS. Latency in seconds; speedup = Base H3 / Sol-H3.
 Base H3: 49 DiT forwards; Sol-H3: 4 DiT forwards.}
\end{table}
\endgroup

Each Sol-H3 latency is the median of five calls after two warmups, with four denoising evaluations verified per call. Timing covers text encoding, denoising, video/audio VAE decoding, and synchronization, excluding model loading, compilation, warmup, and MP4 encoding. Single-GPU execution uses the same SOL/BSA policy and local QKV/output quantization without distributed collectives. It uses the compiled per-tile VAE decoder; multi-GPU execution uses the compiled, globally batched distributed decoder. The Base H3 reference uses 49 denoising evaluations.

As shown in \cref{tab:solh3_scaling}, Sol-H3 achieves 12.73$\times$--16.99$\times$ speedup over Base H3 across the evaluated workloads. On eight GB200 GPUs, five-, ten-, and fifteen-second videos take 1.434\,s, 3.350\,s, and 6.062\,s, respectively. The five-second workload is generated approximately 3.5$\times$ faster than real-time playback.






\subsection{Reference Conditioning and KV Caching}
\label{subsec:reference_kv_exp}


Reference conditioning substantially increases computation overhead. As shown in ~\Cref{tab:reference_kv_50step}(left), one image adds 14,344 effective reference tokens and raises the DiT time to 1.71$\times$ compared with reference-free generation process; one video adds 31,792 tokens for 2.83$\times$, and a video together with an image adds 46,136 tokens for 3.94$\times$. Cost is governed by the reference-token count as the reference materials grows, consistent with the quadratic growth of attention in sequence length. Reference KV caching reduces repeated reference-branch computation, yielding speedups of 1.14$\times$--1.92$\times$, demonstrated in ~\Cref{tab:reference_kv_50step}(right). The benefit grows with the number of references within each modality.  

Although the caching strategy is an approximation method instead of a lossless system optimization, the qualitative comparison demonstrates that this approximation only introduces minor perceptual differences. As shown in \Cref{fig:refcache_quality}, we can see that the generated samples with caching strategy preserve scene structure, subject identity and motion continuity against the full-compute baseline and strictly follow the instruction of video-to-video editing.

\begingroup
\definecolor{refkvdark}{HTML}{386A21}
\definecolor{refkvwash}{HTML}{EFF6E5}
\newcommand{\refkvgain}[1]{\textbf{\textcolor{refkvdark}{#1$\times$}}}
\begin{table}[!t]
\centering
\caption{\textbf{Additional cost introduced by reference tokens and acceleration of reference KV caching.} Left: DiT time relative to
reference-free generation against the effective reference-token count. The latency increases dramatically as the reference materials growing. Right:
effective reference-token counts for every condition and the speedup obtained by reference KV caching strategy, which reducing the computation overhead up to 2$\times$.}
\label{tab:reference_kv_50step}
\begin{minipage}[c]{0.44\linewidth}
\centering
\includegraphics[width=\linewidth]{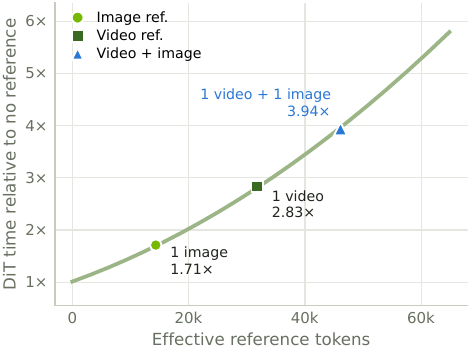}
\end{minipage}%
\hfill
\begin{minipage}[c]{0.53\linewidth}
\centering
\fontsize{8.6}{10.8}\selectfont
\setlength{\tabcolsep}{5pt}
\renewcommand{\arraystretch}{1.30}
\begin{tabular*}{\linewidth}{@{\extracolsep{\fill}}lrr@{}}
\toprule
\textbf{Condition} & \textbf{Eff. ref. tokens} & \textbf{KV cache speedup} \\
\midrule
No reference & 0 & --- \\
\midrule
1 image  & 14,344 & \cellcolor{refkvwash}\refkvgain{1.14} \\
2 images & 28,432 & \cellcolor{refkvwash}\refkvgain{1.35} \\
3 images & 42,776 & \cellcolor{refkvwash}\refkvgain{1.56} \\
\midrule
1 video  & 31,792 & \cellcolor{refkvwash}\refkvgain{1.41} \\
2 videos & 63,584 & \cellcolor{refkvwash}\refkvgain{1.92} \\
\midrule
1 video + 1 image & 46,136 & \cellcolor{refkvwash}\refkvgain{1.65} \\
\bottomrule
\end{tabular*}
\end{minipage}
\end{table}
\endgroup

\begin{figure}[!t]
\centering
\scriptsize
\setlength{\tabcolsep}{2pt}
\renewcommand{\arraystretch}{1.15}
\begin{tabular}{@{}ccc@{}}
\includegraphics[width=0.185\linewidth]{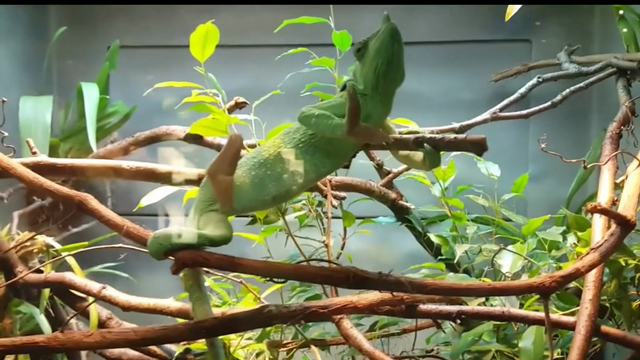} &
\includegraphics[width=0.375\linewidth]{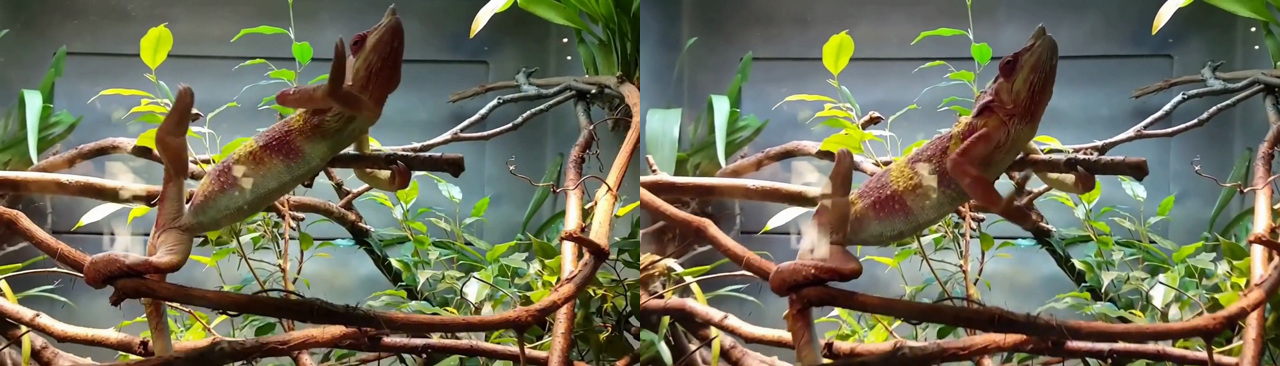} &
\includegraphics[width=0.375\linewidth]{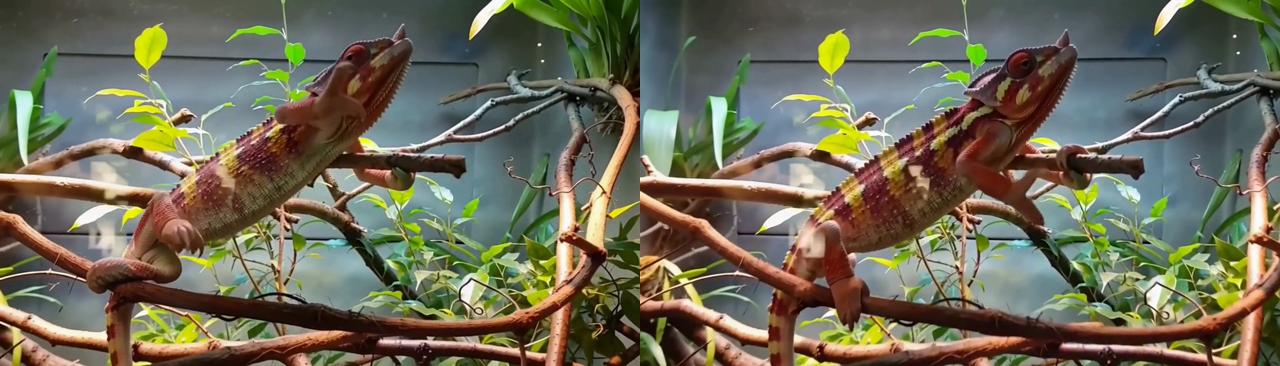} \\
\parbox{0.185\linewidth}{\centering Reference input \\ \emph{Red chameleon crawling on a branch}} &
Teacher (full compute) &
\textbf{Cached reference KV} --- \textbf{\textcolor{nvidiagreen}{1.512$\times$}} \\[6pt]
\includegraphics[width=0.185\linewidth]{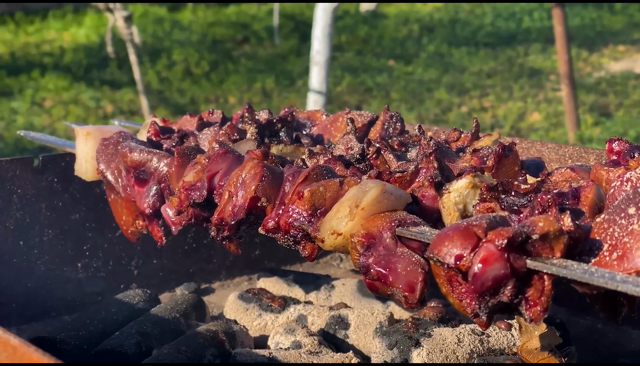} &
\includegraphics[width=0.375\linewidth]{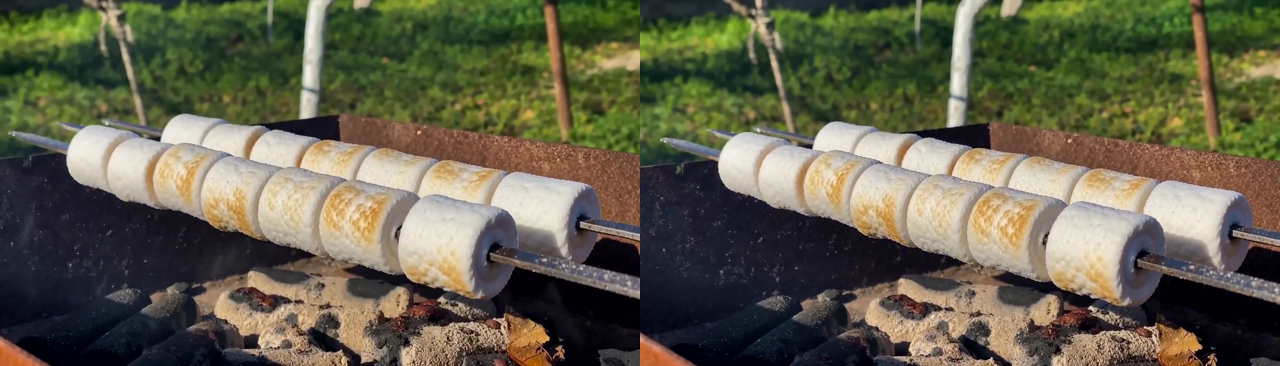} &
\includegraphics[width=0.375\linewidth]{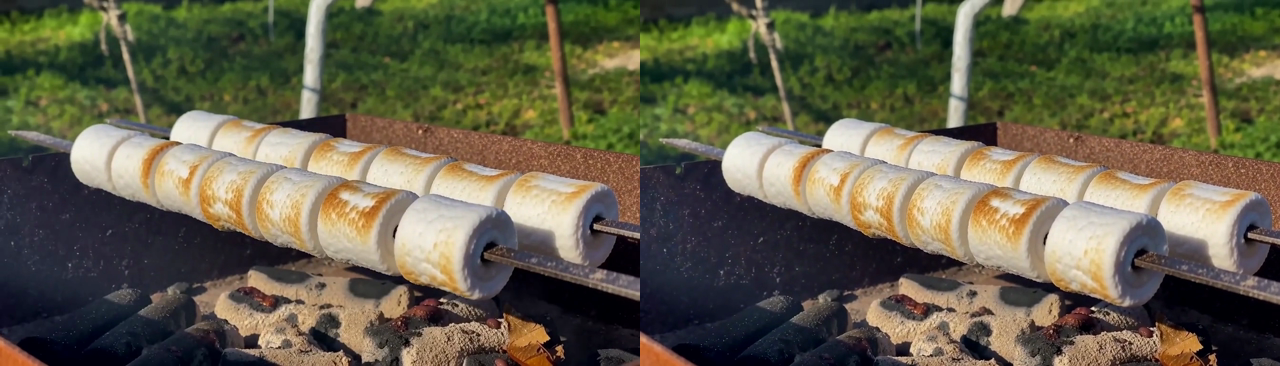} \\
\parbox{0.185\linewidth}{\centering Reference input \\ \emph{Replace meat on skewer with toasted marshmallows}} &
Teacher (full compute) &
\textbf{Cached reference KV} --- \textbf{\textcolor{nvidiagreen}{1.486$\times$}} \\[6pt]
\includegraphics[width=0.185\linewidth]{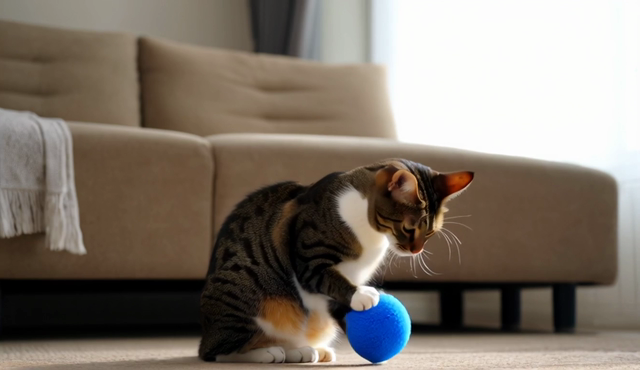} &
\includegraphics[width=0.375\linewidth]{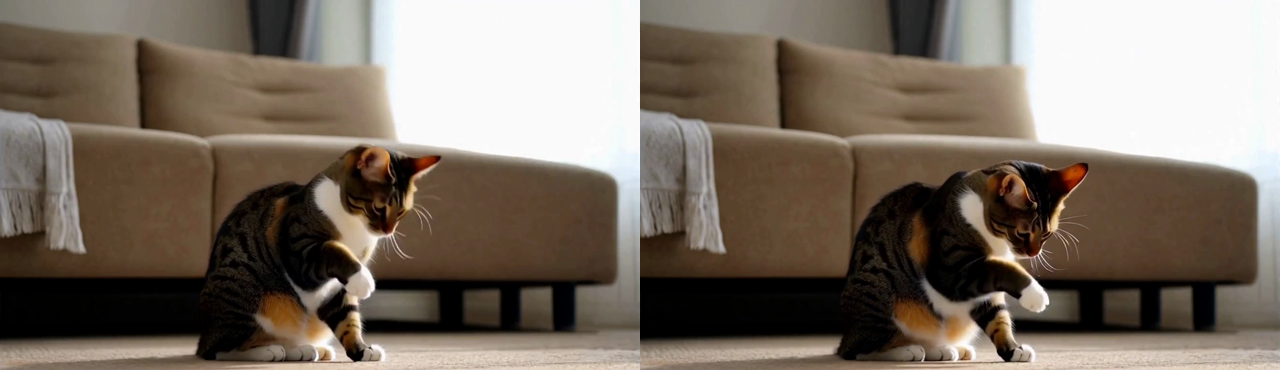} &
\includegraphics[width=0.375\linewidth]{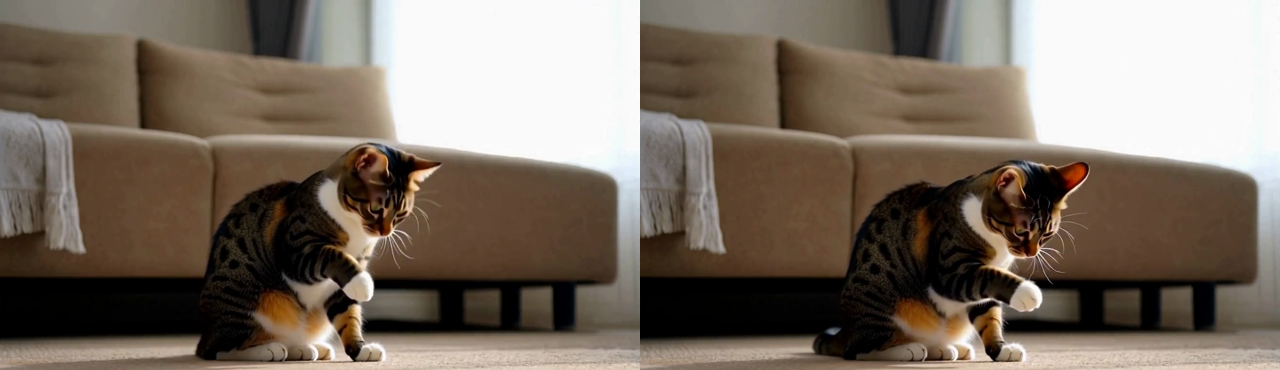} \\
\parbox{0.185\linewidth}{\centering Reference input \\ \emph{Cat pouncing playfully without a toy}} &
Teacher (full compute) &
\textbf{Cached reference KV} --- \textbf{\textcolor{nvidiagreen}{1.416$\times$}} \\
\end{tabular}
\caption{\textbf{Reference KV cache: qualitative comparison.} Each row displays a single frame from the input reference video on the left, two frames from the full-compute teacher baseline in the middle, and two frames from our cached reference KV generation on the right. It demonstrates that this approximation only introduces minor visual differences while preserving visual quality and instruction alignment}
\label{fig:refcache_quality}
\end{figure}

\subsection{Comparison of Few-Step LoRA Adapters}
\label{subsec:fewstep_lora}

We evaluate 120 text-to-video-and-audio (T2VA) tasks and 120 reference-to-video-and-audio (Ref2VA) tasks, spanning diverse subjects, visual styles, actions, and camera movements. Ref2VA includes single- and multiple-reference inputs, direct video editing, and transfer of identity, motion, or camera behavior. Candidates are grouped by task type and number of function evaluations (NFE), with four and six candidates for T2VA at 4 and 8 NFE, respectively, and two and three candidates for Ref2VA. Exhaustive within-group pairing for every task yields 3,000 pairwise comparisons: 720, 1,800, 120, and 360 in these four groups. In addition to the adapters discussed below, the T2VA pool includes FastH3 Preview v1 LoRA~\cite{fastvideo2026fasth3preview} at 4 NFE and the FastH3 V2 full distilled-checkpoint baseline~\cite{fastvideo2026fasth3v2} at 8 NFE. For each task, the compared runs use the frozen rewritten prompt, reference assets, and the same initial video/audio noise.

We combine human assessment on a subset of pairs with VLM assessment of the full 3,000-pair pool, keeping the two sources of judgments separate. Both protocols hide model identities, randomize the A/B presentation, and allow ties. Human raters can play the generated videos with audio and provide an overall preference. For the visual-only VLM assessment, Astra with high reasoning effort receives the generation instruction, 16 uniformly sampled frames from each candidate video, and the available visual reference material; reference videos are likewise sampled into 16 frames. No audio is supplied or scored, and audio-related prompt requirements are excluded from its judgment. Each pair receives one VLM judgment in a fresh context.

Our comparisons favor Larry's Turbo LoRA (v4, step 600)~\cite{larryvrh2026h3turbo} and LightX2V FL2VA Turbo (4-step v1.2)~\cite{lightx2v2026h3turbo} for T2VA at 4 NFE, and Alibaba PAI PDD Acc8~\cite{shaul2026pdd,alibabapai2026h3acc} and the official VDN-H3 stage-dmd-step-250 adapter~\cite{xi2026videodeltanet} at 8 NFE. For Ref2VA, both Alibaba PAI's 4-NFE PDD recipe~\cite{alibabapai2026h3acc} and LightX2V Ref2VA Turbo (v0.1)~\cite{lightx2v2026h3turbo} perform well at 4 NFE, while the public HyperFlow v1.0 release~\cite{videorebirth2026hyperflow} is our preferred choice at 8 NFE.
\subsection{Cross-VAE Latent Translation}
\label{subsec:adapter_exp}

We evaluate the adapter separately from the two-stage generator to isolate conversion cost and reconstruction fidelity. The quality evaluation uses 256 held-out test videos, disjoint from training, development, and earlier evaluation cohorts. Both the predicted latent and the paired LTX teacher latent are decoded with the same frozen Conv Video VAE. We report mean per-video PSNR over the decoded clips and SSIM over 16 uniformly sampled frames. These metrics measure agreement with the LTX reconstruction of the source video, before any Stage-2 denoising.

\begin{table}[H]
\centering
\caption{\textbf{Held-out latent translation quality.} Both checkpoints use the same 194.76M-parameter architecture. Metrics compare decoded adapter predictions with decoded paired LTX teacher latents on the same 256 test clips; they do not include the refiner.}
\label{tab:adapter_quality}
\small
\begin{tabular}{lrrr}
\toprule
\textbf{Checkpoint} & \textbf{Latent MSE $\downarrow$} & \textbf{PSNR (dB) $\uparrow$} & \textbf{SSIM $\uparrow$} \\
\midrule
Earlier decoder-aware checkpoint & 0.11058 & 29.413 & 0.8763 \\
Final checkpoint & \textbf{0.09783} & \textbf{30.618} & \textbf{0.8962} \\
\bottomrule
\end{tabular}
\end{table}

As shown in \cref{tab:adapter_quality}, the final adapter reaches 30.618\,dB PSNR and 0.8962 SSIM. Relative to the earlier checkpoint with the same architecture, the paired PSNR gain is 1.205\,dB, with a 95\% bootstrap interval of $[1.166,1.245]$\,dB; all 256 clips improve in both PSNR and SSIM. The highest-motion quartile gains 1.366\,dB, compared with 1.059\,dB in the lowest-motion quartile. This improvement comes from continued decoder-aware training with a larger learning rate, batch size, and exposure budget, without increasing inference-time model size. It does not isolate the contribution of each training change.

\begin{table}[H]
\centering
\caption{\textbf{Isolated H3-to-LTX conversion cost.} One H100 80GB, BF16, batch one, 192 source frames at $1344\times768$. Latency is the median synchronized wall time after warmup; memory is peak allocated device memory for each conversion arm. The adapter includes geometry alignment but excludes the separate $\times2$ upscaler.}
\label{tab:handoff}
\small
\begin{tabular}{lrrrr}
\toprule
\textbf{Conversion} & \textbf{Params (M)} & \textbf{Latency (ms)} & \textbf{Peak (GiB)} & \textbf{Speedup} \\
\midrule
Full VAE round trip & 2,742.47 & 12,748.6 & 13.788 & $1.0\times$ \\
Tiny AutoEncoder round trip & 23.11 & 201.9 & 23.846 & $63.1\times$ \\
Latent adapter & 194.76 & \textbf{63.1} & \textbf{0.785} & $\mathbf{202.0\times}$ \\
\bottomrule
\end{tabular}
\end{table}

For conversion cost, \cref{tab:handoff} profiles a batch-one, 192-frame clip at $1344\times768$ on one H100 80GB in BF16. The adapter, including its fixed geometry transform, takes 63.1\,ms versus 12,748.6\,ms for H3 decoding followed by LTX encoding, a $202.0\times$ reduction in median conversion latency. Peak allocated memory falls from 13.79 to 0.785\,GiB. A Tiny AutoEncoder round trip takes 201.9\,ms under the same input geometry. This profile uses an earlier checkpoint of the same 194.76M-parameter architecture, whereas \cref{tab:adapter_quality} evaluates the final trained weights. The measurements exclude spatial upsampling, denoising, final video decoding, and audio processing; they therefore quantify the cost of the handoff module, not an end-to-end speedup or a quality-equivalent replacement for the full VAE round trip.

\subsection{Precision-Preserving LoRA Fusion}
\label{sec:exp_lora_qualitative}

\paragraph{Numerical preservation and fusion cost.}
We first verify that consumer fusion preserves the selected adapter's computation. With the evaluated eight-step adapter and BF16 linear compute, consumer fusion matches native separate branches byte for byte on three 15-second samples, across all eight denoising evaluations and four GPU ranks, including the final raw video and audio tensors. Both modes retain the same other acceleration settings, including sparse attention and low-precision attention transport. This establishes agreement for the tested inputs and software environment, rather than equivalence to an entirely unaccelerated model.

\Cref{tab:lora_fusion} compares the three execution modes on four GB200 GPUs for one $1344\times768$, 362-frame workload. Native and fused branches each use five interleaved warm measurements; merged weights use three warm measurements in a separate process on the same GPU allocation. Consumer fusion reduces median pipeline time by 0.694\,s (2.64\%), recovering 46.3\% of the native branches' additional time relative to merged weights. It remains 0.804\,s (3.24\%) slower than weight merging, which produces different numerical results. Timing includes text encoding, denoising, audio/video decoding, and synchronization, while excluding model loading, warmup, and MP4 encoding. The saving includes the fused FFN gate and residual operations; the LoRA GEMMs remain unchanged.

\begin{table}[H]
\centering
\caption{\textbf{LoRA precision and execution cost.} Median pipeline latency for eight denoising evaluations on four GB200 GPUs. Numerical agreement is relative to native separate branches under the same other acceleration settings. Merged weights provide a speed reference with different outputs.}
\label{tab:lora_fusion}
\small
\begin{tabular}{lrc}
\toprule
\textbf{LoRA execution} & \textbf{Time (s) $\downarrow$} & \textbf{Bitwise agreement} \\
\midrule
Merged BF16 weights & 24.785 & No \\
Native separate branches & 26.283 & Reference \\
Consumer fusion & 25.589 & Yes (tested inputs) \\
\bottomrule
\end{tabular}
\end{table}

\begin{figure}[H]
\centering
\scriptsize
\setlength{\tabcolsep}{2pt}
\renewcommand{\arraystretch}{1.05}
\begin{tabular}{@{}ccc@{}}
\textbf{Merged BF16 weights} & \textbf{Native separate branches} & \textbf{Consumer fusion} \\
8 steps & 8 steps & 8 steps \\[3pt]
\multicolumn{3}{@{}l@{}}{\textit{7.50\,s (frame 180)}} \\[1pt]
\includegraphics[width=0.325\linewidth]{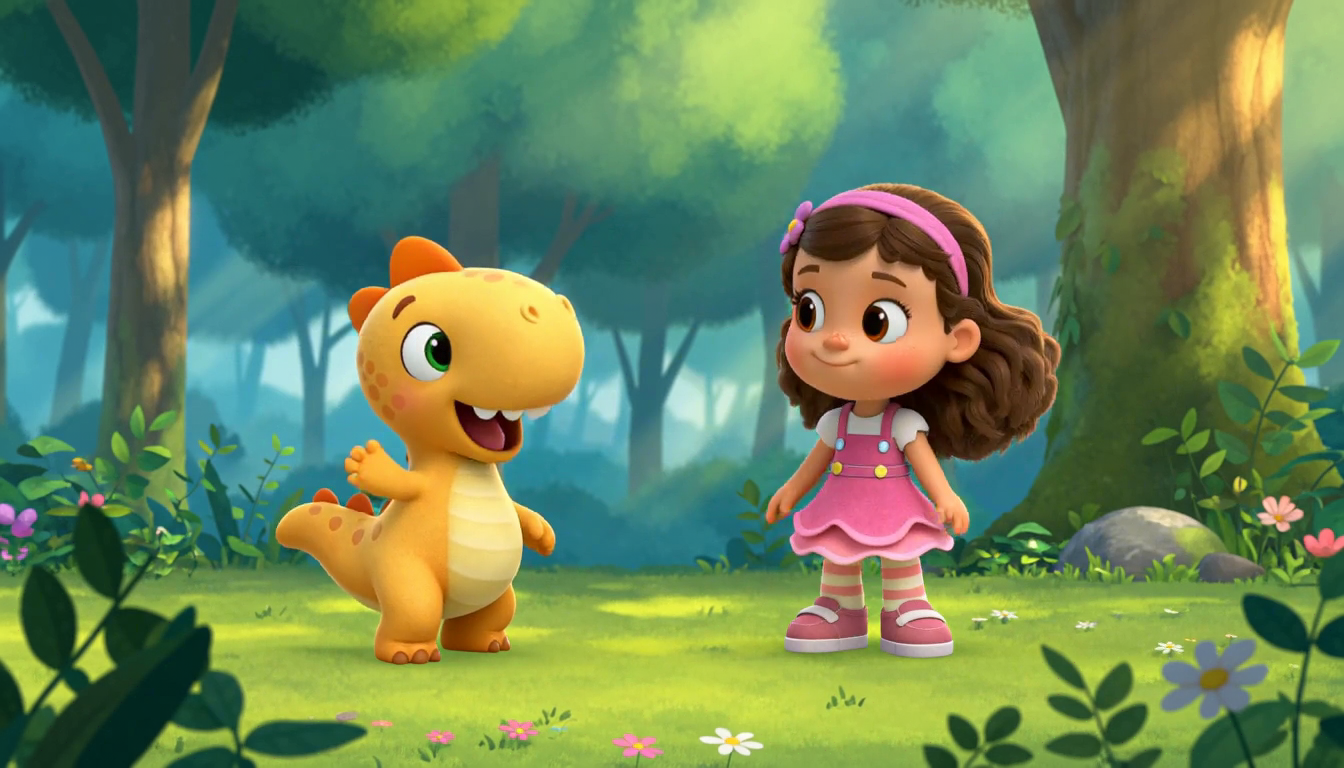} &
\includegraphics[width=0.325\linewidth]{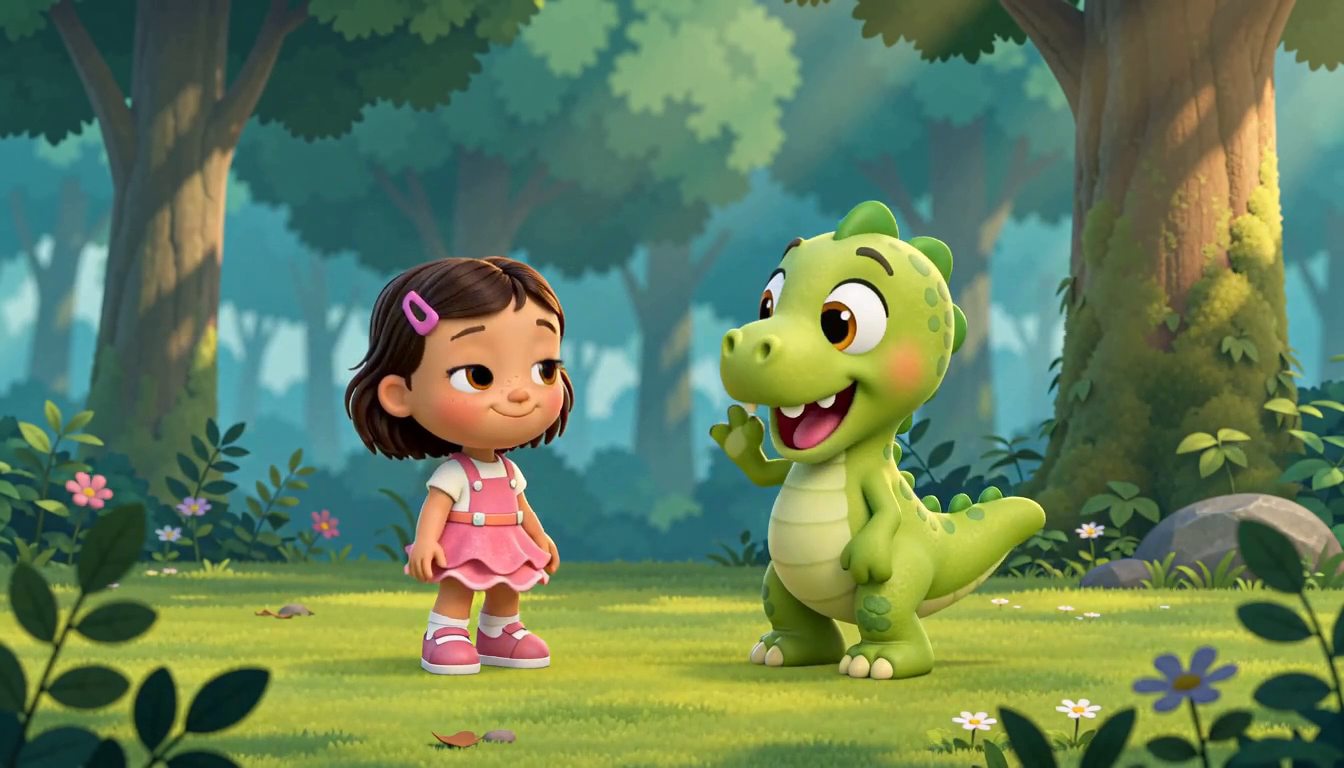} &
\includegraphics[width=0.325\linewidth]{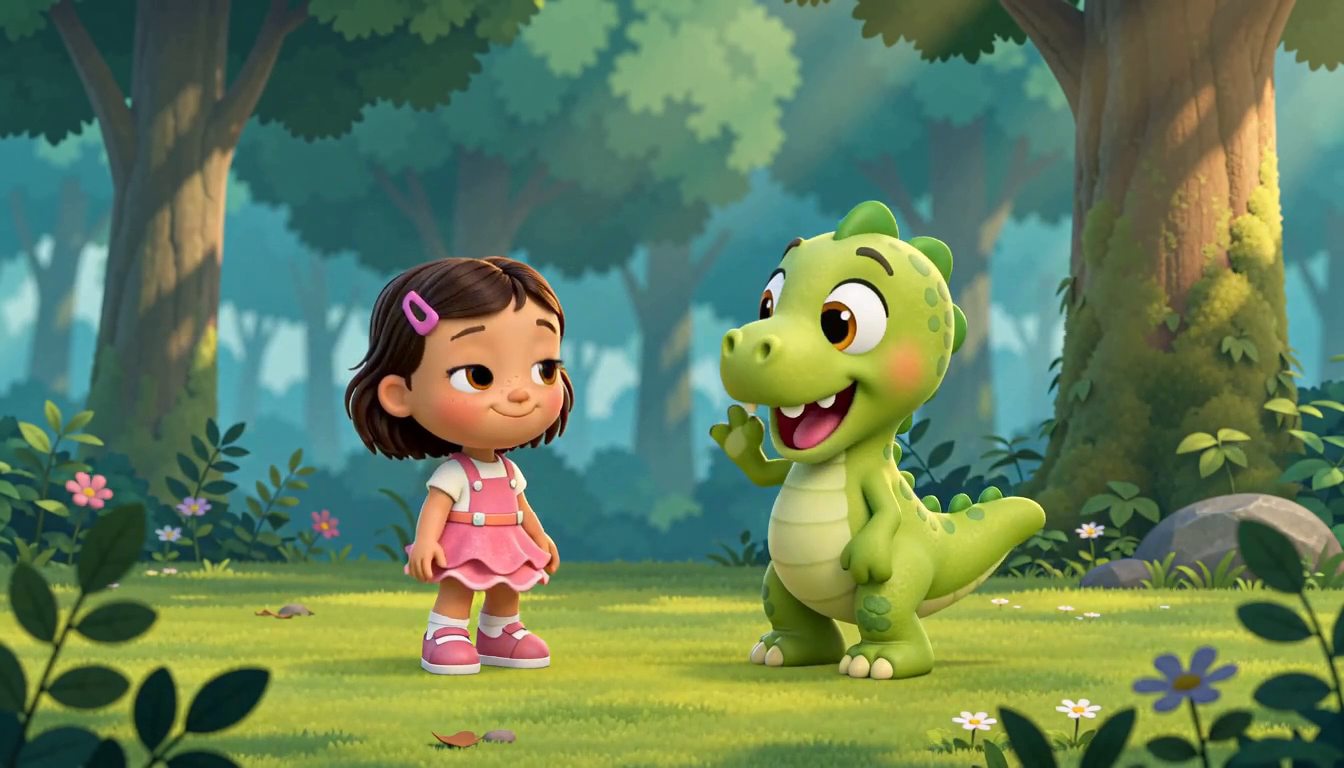} \\[4pt]
\multicolumn{3}{@{}l@{}}{\textit{12.50\,s (frame 300)}} \\[1pt]
\includegraphics[width=0.325\linewidth]{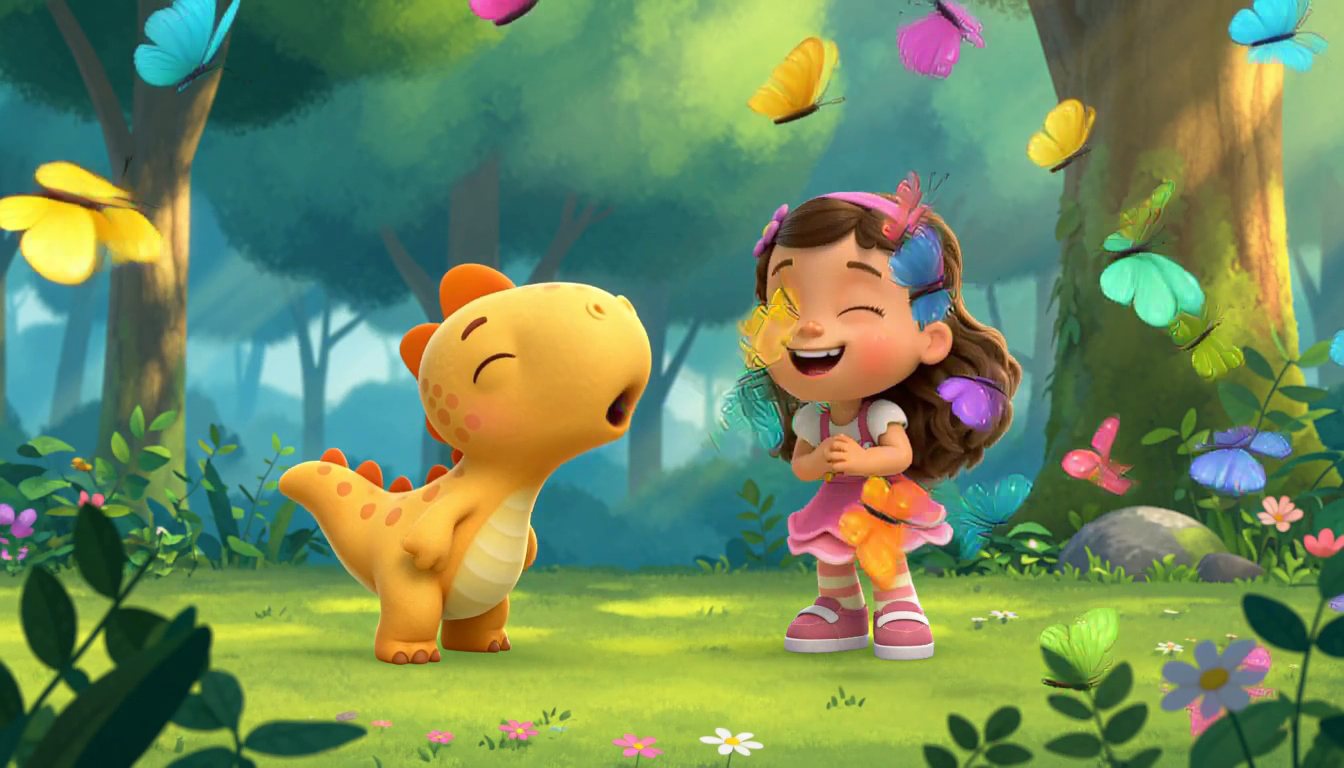} &
\includegraphics[width=0.325\linewidth]{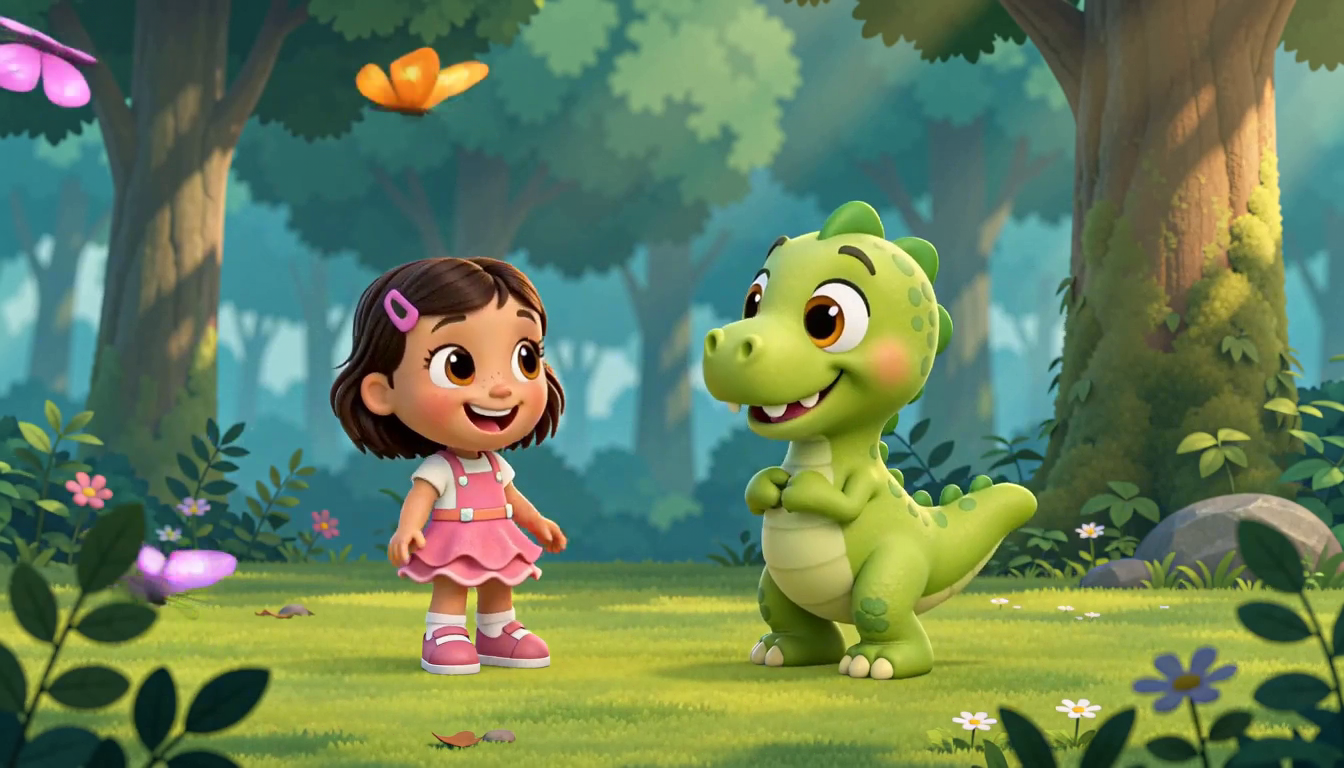} &
\includegraphics[width=0.325\linewidth]{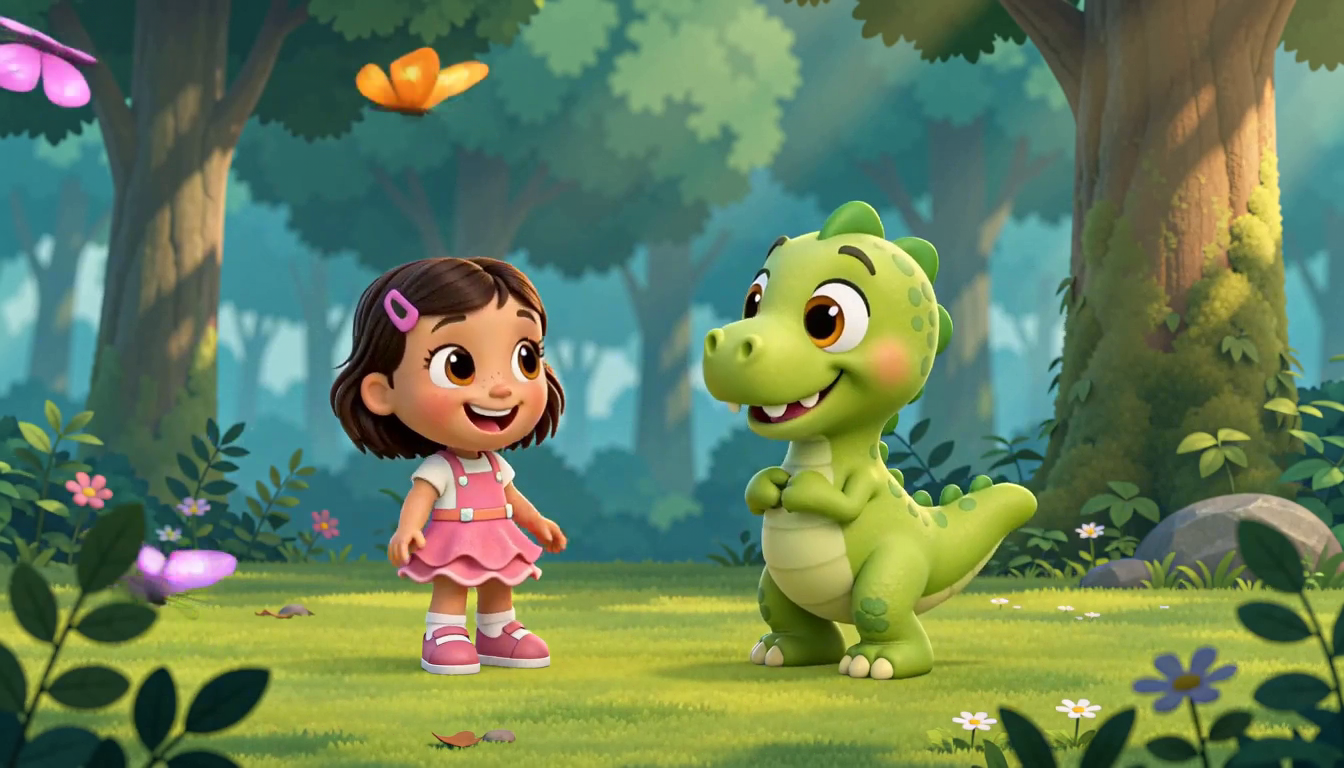} \\[5pt]
\multicolumn{3}{@{}l@{}}{\textit{Butterfly detail (event-aligned crops; times differ)}} \\[1pt]
\includegraphics[width=0.325\linewidth]{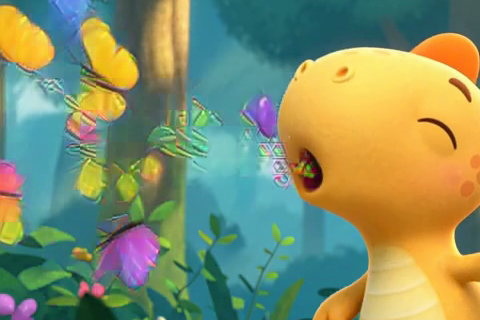} &
\includegraphics[width=0.325\linewidth]{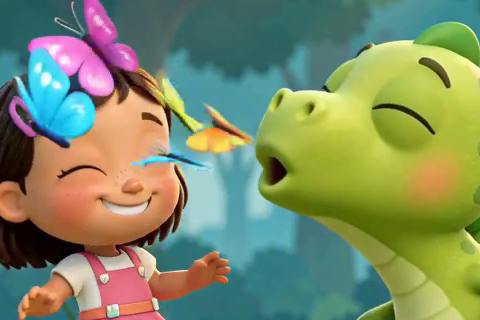} &
\includegraphics[width=0.325\linewidth]{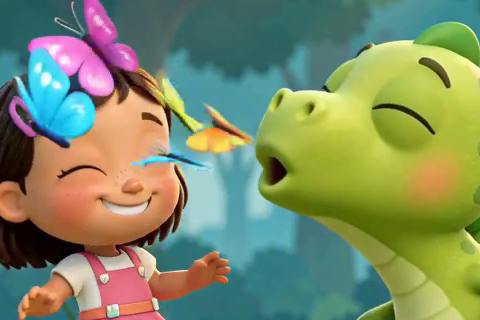} \\
10.50\,s & 11.33\,s & 11.33\,s \\
\end{tabular}
\caption{\textbf{LoRA execution changes the generation trajectory.} The same eight-step checkpoint, prompt, seed, and initial noise are evaluated on four GB200 GPUs at $1344\times768$, with the other acceleration settings fixed. The upper rows compare identical timestamps; the bottom row magnifies the butterfly burst at the indicated times. All images come from lossless frames exported before video encoding. Native branches and consumer fusion match byte for byte in the decoded video and audio; merging the weights changes the scene and introduces grid-like patterns around the butterflies in this example.}
\label{fig:lora_compare}
\end{figure}

\paragraph{Qualitative effect of weight merging.}
We isolate the execution mode using a fixed eight-step adapter and a 15-second animated scene in which a girl meets a dinosaur that sneezes butterflies. The three arms in \cref{fig:lora_compare} share the prompt, seed, actual initial noise, conditioning, sampling schedule, and other runtime optimizations. Weight merging changes the dinosaur's appearance, the girl's hairstyle, their relative positions, and the timing of the butterfly burst. The magnified view also shows grid-like color patterns around the butterflies in the merged-weight output, visible in the decoded frames before video encoding. Native branches and consumer fusion preserve the same generated scene: the rerun verifies byte agreement at every denoising evaluation on all four ranks, in the final decoded video and audio, and in all 362 exported frames. This example illustrates how a numerical change can alter a generation trajectory; it is not a comparison between learned adapters or a quantitative perceptual-quality evaluation.

\section{Discussion and Conclusion}
\label{sec:conclusion}

In this report, we present Sol-H3, a full-stack efficient inference pipeline for the MiniMax-H3 video-audio generator. By combining a cross-resolution two-stage generation pipeline with a kernel-level optimization stack discovered via Recursive Self-Improvement (RSI), Sol-H3 achieves unprecedented generation speeds across both cloud and edge environments. 

A key takeaway from the development of Sol-H3 is the complementary relationship between human architectural insight and automated optimization. In our work, the macro-level inference framework is fundamentally driven by human design, whereas the micro-level implementation optimizations are systematically executed by the RSI loop. 

Specifically, RSI is deployed to handle execution bottlenecks where the optimization space is vast but the outcomes are strictly verifiable (e.g., maintaining mathematical correctness). Its primary contributions in this pipeline include:
\begin{itemize}[leftmargin=1.5em, itemsep=2pt]
    \item \textbf{Communication collectives and quantization:} Jointly searching the configuration space of data compression and network transfer to eliminate distributed inference bottlenecks.
    \item \textbf{Fusion kernel implementation:} Automatically discovering and generating optimal operator groupings to minimize redundant HBM traffic.
    \item \textbf{Sparse attention implementation:} Engineering highly efficient, hardware-aware attention mechanisms to manage massive sequence lengths.
    \item \textbf{Precision-preserving LoRA integration:} Ensuring that low-rank adaptation operations are accelerated without compromising numerical fidelity or generation quality.
\end{itemize}

In contrast, the open-ended, inherently lossy, and architectural components of the pipeline require a level of paradigm-shifting intuition that RSI currently cannot provide. These human-designed components include:
\begin{itemize}[leftmargin=1.5em, itemsep=2pt]
    \item \textbf{The cross-resolution two-stage framework:} Conceptualizing the overarching algorithmic strategy to exploit the step-wise nature of diffusion models, balancing early global layout generation with later high-resolution refinement.
    \item \textbf{The latent adapter:} Recognizing the system-level trade-offs of standard VAE decoders/encoders, and designing a custom adapter to efficiently bridge the coarse-to-fine generation stages at the cost of strict scene-identity invariance.
\end{itemize}

Based on this dichotomy, we observe that RSI is exceptionally well-suited for systematically resolving deterministic, well-bounded engineering problems where success can be mathematically verified against a baseline. However, for open-ended, architectural pipeline designs that necessitate lossy approximations or holistic trade-offs, human engineering remains indispensable. Ultimately, this synergy of human algorithmic design and RSI-driven kernel optimization enables faster-than-real-time high-resolution video synthesis, drastically lowering the barrier to deploying massive video generation models in production.

\newpage
\appendix
\onecolumn
\begin{center}\Large\bfseries Appendix\end{center}
\vspace{4pt}
\section{Latent Adapter Training and Evaluation}
\label{app:adapter}

\paragraph{Paired data and geometry.}
The source corpus contains 80,000 videos stored in eight archives. Each source video is resized identically before encoding with the frozen H3 and LTX-2.5 Conv VAEs. We use deterministic posterior modes and the released per-channel normalization statistics. For a 192-frame clip at $1344\times768$, the H3 tensor has shape $24\times57\times48\times84$ in channel--time--height--width order. LTX pads the input to 193 frames and produces $128\times25\times24\times42$. H3 token positions follow the 17-frame chunking rule, with anchors $0,4,8,12,16,17,21,\ldots$ and the configured removal of three trailing tokens. The fixed alignment in \cref{subsec:adapter} yields $384\times25\times24\times42$ features, retaining each H3 token in one of three temporal slots before spatial pixel-unshuffle. The skip projection maps these features to 128 channels and remains fixed after ridge-regression initialization; the residual network is trained end to end.

\paragraph{Optimization.}
The final decoder-aware optimization starts from an earlier trained adapter and uses 65,536 cached video pairs at native resolution. Two phases of 2,500 and 5,000 updates use global batch size 32, for 240,000 additional sample presentations. Each phase uses AdamW with peak learning rate $8\times10^{-5}$, $\beta=(0.9,0.95)$, zero weight decay, and gradient clipping at 1.0. The respective warmup lengths are 100 and 150 updates, followed by cosine decay to 5\% of the peak learning rate; the second phase restarts the optimizer from the first phase's weights. Training uses BF16 autocast and activation checkpointing through the residual stack and the frozen decoder. The objective in \cref{eq:adapter_loss} is evaluated on the full spatial grid, with decoded-video supervision weight $\alpha=160$. Only the adapter's residual path is updated; the affine skip and both VAEs remain fixed. The reported final checkpoint uses the trained weights directly, without EMA averaging.

\paragraph{Held-out evaluation.}
Sample identifiers are assigned to train, validation, and test partitions by a stable hash with a 98/1/1 split. The final development and test cohorts contain 256 videos each and exclude identifiers used in earlier evaluation cohorts. Checkpoint selection uses development data. In the final evaluator, predictions and teacher reconstructions are clamped to $[0,1]$ and compared over their common decoded temporal extent, including the VAE padding rather than an explicit crop back to 192 frames. PSNR is computed per video and then averaged across videos; SSIM uses 16 uniformly sampled decoded frames. Motion quartiles are defined by temporal differences in the teacher latents. Paired confidence intervals use 10,000 bootstrap resamples over the same clip identifiers. These reconstruction measurements do not evaluate prompt adherence, audio quality, or the output of subsequent LTX refinement.

\paragraph{Conversion benchmark.}
All three arms in \cref{tab:handoff} consume the same cached H3 latent and output normalized LTX latents at the same pixel resolution. The full round trip uses the released H3 decoder and tiled LTX Conv encoder. The Tiny AutoEncoder arm uses TAEH3 decoding and TAELTX encoding with their parallel execution mode. Both round-trip arms use one warmup and three measured iterations; the adapter uses three warmups and 20 measured iterations. CUDA synchronization brackets wall-time measurements, and model loading and disk I/O are excluded. The profiled adapter is an earlier checkpoint with the same geometry, width, depth, and parameter count as the final quality checkpoint. Peak allocation is specific to each implementation and excludes the co-resident generative pipeline; it is not an estimate of total Spark memory usage.

\section{LoRA Precision Diagnosis and Fusion}
\label{app:lora_fusion}

\paragraph{Separating weight rounding from GEMM precision.}
The merge diagnostic fixes the checkpoint, prompt, initial noise and sampling schedule, and compares the first denoising evaluation on four GB200 GPUs with the other runtime optimizations disabled. With linear computation held in FP32, merging into BF16 weights gives 35.26\% relative RMSE in the video prediction against the original separate-branch baseline; retaining the merged weights in FP32 reduces this to 6.24\%. FP32 GEMMs alone therefore do not recover updates already lost during weight storage. Changing the separate branches themselves to FP32 gives 6.08\% relative RMSE against the same baseline, showing why higher precision is also insufficient to reproduce the original mixed-precision computation exactly. These are prediction discrepancies for one input at one denoising evaluation, not perceptual-quality scores. The six-projection counts in \cref{subsec:fusion} use FP32 reconstructions of $W+BA$ and count nonzero entries of $BA$ for which casting the sum to BF16 returns the original $W$ entry.

\paragraph{Implementation scope.}
The released implementation\footnote{\href{https://github.com/NVlabs/Sana/pull/503}{NVlabs/Sana PR~\#503}, commit \texttt{51b14fd6f262}.} accepts the supported public PEFT LoRA~\cite{hu2022lora} and FastVideo v2 hybrid formats. Hybrid \texttt{.diff}/\texttt{.diff\_b} corrections are applied to the base parameters with FP32 accumulation before conversion to the parameter dtype; low-rank factors remain separate. Consumer fusion requires inference mode, one active scale-1 adapter, ordinary BF16 linear layers, identity LoRA dropout, and no projection hooks. Unsupported installation configurations are rejected before consumer patches are installed; unsupported runtime states fall back to native PEFT. QKV consumer fusion uses the multi-GPU Ulysses path, while single-GPU dense Q/K/V projections and projections outside the main transformer blocks retain native execution.

\paragraph{Execution modes.}
The public runtime defaults to \texttt{merged}. The \texttt{separate} and \texttt{fused} modes retain LoRA branches and require \texttt{compute-quant=none}; the latter enables the consumer kernels. This disables quantized linear compute, not the independent attention or communication optimizations. The controlled comparison in \cref{tab:lora_fusion} explicitly selects each mode; enabling consumer fusion is not implicit in every pipeline measurement. The released regression suite covers BF16 rounding boundaries, non-contiguous inputs, indexed gates, promoted-dtype fallback, BF16/INT8 QKV packing, and row addressing beyond $2^{31}$ elements. Full-pipeline byte agreement remains scoped to the tested inputs and their recorded GPU and software configurations.

\paragraph{Qualitative case study.}
The additional rerun in \cref{fig:lora_compare} uses seed 42, eight denoising evaluations, four GB200 GPUs, and 362 frames at $1344\times768$ and 24\,FPS. Each arm performs its natural random draws and verifies the resulting tensors and generator states against the same stored reference before replay; initial conditioning and the video/audio sampling grids are also checked. The native and fused arms match in every recorded prediction and in the complete decoded video and audio. We export all frames directly to lossless 8-bit RGB PNGs before MP4 encoding and compare their pixel hashes. The full-frame rows use common zero-based indices 180 and 300. The $480\times320$ detail crops instead follow the butterfly burst at the explicitly labeled times, because merging changes its timing as well as its appearance. These audited runs include tensor hashing and frame export and are separate from the warm latency measurements in \cref{tab:lora_fusion}.

\clearpage
{
  \small
  \bibliographystyle{unsrtnat}
  \bibliography{ref}
}

\end{document}